%% file: main.tex
\PassOptionsToPackage{table,xcdraw}{xcolor}

\documentclass[]{gradient}

\usepackage{array}
\usepackage{subcaption}

\tcbuselibrary{listings}

\usepackage{amsmath}
\usepackage{amssymb}
\usepackage{mathtools}

\usepackage{natbib}

\usepackage{wrapfig}

\usepackage{algorithm}
\usepackage{algorithmic}
\usepackage{enumitem}
\usepackage{xspace}
\usepackage{float}

\input{macro}

\title{AOI: Turning Failed Trajectories into Training Signals for Autonomous Cloud Diagnosis}

\author{Pei Yang\textsuperscript{1*}
    \quad Wanyi Chen\textsuperscript{2*}
    \quad Asuka Yuxi Zheng\textsuperscript{3*}
    \quad Xueqian Li\textsuperscript{4}
    \quad Xiang Li\textsuperscript{5}
    \quad Haoqin Tu\textsuperscript{3}
    \quad Jie Xiao\textsuperscript{1}
    \quad Yifan Pang\textsuperscript{6}
    \quad Dongdong Zhang\textsuperscript{7}
    \quad Fuqiang Li\textsuperscript{8}
    \quad Alfred Long\textsuperscript{1}
    \quad Lynn Ai\textsuperscript{1}
    \quad Eric Yang\textsuperscript{1}
    \quad Bill Shi\textsuperscript{1\dag}
}

\affiliation{
    \textsuperscript{1}Gradient \quad
    \textsuperscript{2}Soochow University \quad
    \textsuperscript{3}UC Santa Cruz \quad
    \textsuperscript{4}Georgia Institute of Technology \\
    \textsuperscript{5}University College London \quad
    \textsuperscript{6}Cookiy.ai \quad
    \textsuperscript{7}WeJoy \quad
    \textsuperscript{8}ByteDance \\
    \textsuperscript{*}Equal contribution \quad \textsuperscript{\dag}Corresponding author
}

\date{Mar 16, 2026}
\correspondence{tianyu@gradient.network}
\sourcecode{https://github.com/OpenEdgeHQ/aoi}

\abstract{Large language model (LLM) agents offer a promising data-driven approach to automating Site Reliability Engineering (SRE), yet their enterprise deployment is fundamentally constrained by three challenges: restricted access to proprietary operational data, unsafe action execution under permission-governed environments, and the inability of closed systems to improve from failure trajectories.

We present \method (Autonomous Operations Intelligence), a trainable multi-agent framework that formulates automated operations as a structured trajectory learning problem under security constraints.
Our approach integrates three key components.
First, we introduce a \textbf{trainable diagnostic system} that applies Group Relative Policy Optimization (GRPO) to distill expert-level operational knowledge into locally deployed open-source models, enabling preference-based learning without exposing sensitive data.
Second, we design a \textbf{read--write separated execution architecture} that decomposes operational trajectories into observation, reasoning, and action phases, allowing safe learning over diagnostic traces while preventing unauthorized state mutation.
Third, we propose a \textbf{Failure Trajectory Closed-Loop Evolver} that mines unsuccessful diagnostic trajectories and converts them into corrective supervision signals, enabling continual data augmentation and distributional refinement within a closed environment.

Evaluated on the \benchmark benchmark, our contributions yield
  cumulative gains across three evaluation dimensions.
  (1)~The \method runtime alone---without any task-specific
  training---achieves \textbf{66.3\%} best@5 success on all 86 tasks,
  outperforming the prior state-of-the-art (41.9\%) by
  \textbf{24.4 percentage points}.
  (2)~Adding Observer GRPO training, a locally deployed 14B model
  reaches \textbf{42.9\%} avg@1 on 63 held-out tasks with unseen
  fault types, surpassing Claude Sonnet 4.5 (41.3\%) without
  multi-run sampling.
  (3)~The Evolver further converts 37 previously failed trajectories
  into diagnostic guidance, improving end-to-end avg@5 by
  \textbf{4.8 percentage points} while reducing run-to-run variance
  by \textbf{35\%}.}

\begin{document}

\maketitle

\begin{figure*}[t]
  \centering
  \includegraphics[width=\textwidth]{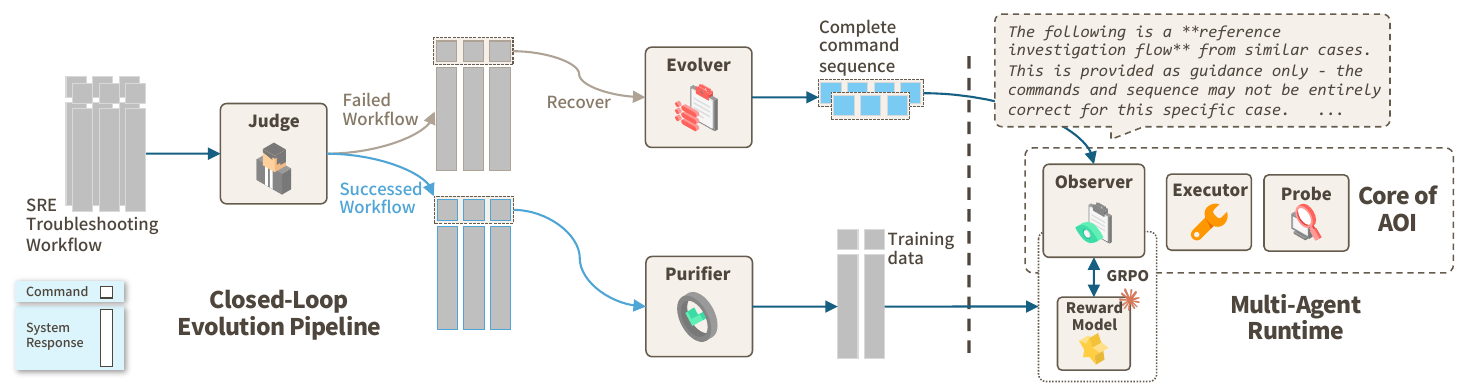}
  \caption{\textbf{AOI System Overview.} Left: \emph{Closed-Loop Evolution Pipeline}---a Judge classifies SRE troubleshooting workflows by outcome. Failed workflows are repaired by the Evolver into corrected command sequences that serve as diagnostic guidance at inference time. Successful workflows are distilled by the Purifier into optimal diagnostic paths that serve as training data. Right: \emph{Multi-Agent Runtime}---the Observer coordinates read-only diagnosis (Probe) and write-gated remediation (Executor). The Observer is trained via GRPO, and at inference time receives the Evolver's corrected plans as structured prompts.}
  \label{fig:overview}
\end{figure*}

\input{section/01-security_introduction}
\input{section/02-related}
\input{section/03-05-security_Approach}
\input{section/06-experiments}
\input{section/07-discussion}
\input{section/08-limitations}
\input{section/09-conclusion}

\clearpage

\bibliography{references}

\newpage
\appendix
\input{section/10-appendix-revised-feb8}

\end{document}

%% file: macro.tex
\graphicspath{{figures/}}

\definecolor{colorA}{RGB}{162,206,222}
\definecolor{colorB}{RGB}{242,189,109}
\definecolor{colorC}{RGB}{124,152,133}
\definecolor{colorD}{RGB}{196,191,186}
\definecolor{promptboxlightgray}{gray}{0.95}
\definecolor{promptboxblue}{RGB}{70,130,180}

\newcommand{\method}{AOI\xspace}

\newcommand{\benchmark}{AIOpsLab\xspace}


%% file: section/01-security_introduction.tex
\section{Introduction}
Site Reliability Engineering (SRE) has become the backbone of modern digital infrastructure, ensuring the stability that users rely on. As these systems grow in complexity, the drive for efficiency naturally accelerates, pushing teams toward automation to minimize downtime~\cite{sre_book}. Large Language Model (LLM) agents offer a promising path toward automation, as recent work demonstrates that LLMs can use tools~\cite{toolformer, weng_agent}, execute multi-step plans via chain-of-thought reasoning~\cite{cot, react}, and interact with complex system environments~\cite{aiopslab, stratus}. However, in enterprise deployments, the security of the agent becomes the defining factor for its adoption, which includes data privacy, permission boundaries, and execution safety. Based on these two, we phrase them as challenges that hold back autonomous SRE agents in real-world enterprise operating systems.

Deploying agents in enterprise environments presents a dual challenge of security and adaptability. (1) Strictly governed operational protocols demand a clear separation between diagnostic (or ``read'') and remediation (or ``write'') privileges of the system, yet standard automation often conflates these rights, proposing the need of a system that enforces granular execution permissions only when strictly necessary. (2) Compounding this complexity, data privacy specifically in the SRE environment mandates force reliance on smaller, locally deployed models ({$<$}100B)\footnote{A 100B-parameter model under FP16 requires ${\sim}$200\,GB for weights alone; adding KV cache for 14K-token context (${\sim}$10\,GB) and system overhead brings the total to ${\sim}$215\,GB, far exceeding the deployment cost budget of most enterprises.} that lack expert reasoning; these constrained systems remain fragile and static, failing to take advantage of valuable learning signals from diagnostic failures to adapt to evolving cloud environments.


To meet with these requirements, we propose \textbf{\method} (Autonomous Operations Intelligence), designed to address both challenges through two integrated mechanisms. (1) To address the first challenge that current systems mix the ``read'' from ``write'' permissions in the autonomous setup, instead of deploying simple prompt engineering on a single LLM, we introduce a multi-agent system~\cite{autogen, metagpt}comprising three parts: Observer, Probe, and Executor that strictly separates the ``read'' and ``write'' actions. 
In detail, high-risk ``write'' commands are technically isolated and can only be triggered after sufficient evidence is gathered and verified, aligning the agent's behavior with the principle of least privilege~\cite{least_privilege}.
(2) Building upon this agent architecture, we propose a trainable and evolving system designed to elevate the performance of our AOI beyond the limitations of static deployments. 
Specifically, we employ Group Relative Policy Optimization (GRPO)~\cite{grpo} to fine-tune the Observer component (i.e., a 14B open-weight LLM), distilling expert-level diagnostic capabilities~\cite{distillation} that bridge the gap between small-scale models and proprietary alternatives. 
To facilitate continuous learning during model inference, we introduce an Evolver dedicated to Failure Trajectory collection, which allows the Observer to systematically learn from its own operational experiences. 
This internal feedback mechanism serves to recycle failed diagnostic trajectories into high-quality corrective training signals, fostering a self-sustaining cycle where the agent continuously evolves. 
By doing so, the system progressively improves its diagnostic precision through real-world interactions while maintaining the strict data isolation required by real environments.
By leveraging our agentic architecture and training pipeline, the 14B model approaches Claude Sonnet 4.5 on the \benchmark benchmark. This result validates the effectiveness of our AOI in enabling small, locally deployable models to close the gap with frontier models.

Extensive empirical evaluations on the \benchmark benchmark substantiate the efficacy of our design. First, the \method runtime alone---without any task-specific training---achieves \textbf{+24.4\%} higher success rate in the best@5 setting compared with the prior state-of-the-art (41.9\% vs.\ 66.3\%), demonstrating that read-write separation yields immediate architectural gains.
Second, Observer GRPO training on only 23 tasks generalizes to 63 held-out tasks spanning unseen fault types, lifting avg@1 from 33.7\% to \textbf{42.9\%}---surpassing Claude Sonnet 4.5 (41.3\%) without multi-run sampling.
Third, the Evolver converts all 37 previously failed trajectories (43\% of the benchmark) into structural diagnostic prompts, improving end-to-end avg@5 by \textbf{4.8\%} while reducing run-to-run variance by \textbf{35\%}.
These results confirm that enforcing strict security constraints and closing the data loop not only ensures safety but also drives robust, reproducible operational capability.

    
    

%% file: section/02-related.tex
\section{Related Work}
\label{sec:related}

\textbf{LLM Agents for AIOps.}
The application of LLMs to cloud operations spans log analysis~\cite{logbert}, anomaly detection~\cite{aiops_survey}, root cause analysis~\cite{rca_survey, rcacopilot}, and interactive diagnosis. STRATUS~\cite{stratus} employs multi-agent collaboration but tightly couples reasoning with execution, causing safety issues and brittle long-horizon behavior. Concurrent work explores retrieval-augmented diagnosis~\cite{rag} and chain-of-thought prompting~\cite{cot} for RCA. \method differs by \emph{architecturally enforcing} safety through role separation rather than relying on prompt-based guardrails.

\noindent \textbf{Safe Agentic Systems.}
Safety mechanisms for LLM agents include action filtering~\cite{toolformer, weng_agent}, sandboxed execution environments~\cite{swe_bench, react}, and reinforcement learning from human feedback~\cite{instructgpt}. These approaches treat safety as an add-on constraint. We instead design safety into the system architecture: the Observer \emph{cannot} directly execute commands; the Probe \emph{cannot} mutate state; the Executor operates under strict whitelists. This separation-of-concerns approach is inspired by classic operating system security principles~\cite{least_privilege}.

\noindent \textbf{Learning from Failures.}
Failure analysis has long been central to software engineering~\cite{debug}. Recent work uses failed test cases to guide program repair~\cite{apr}, while Reflexion~\cite{reflexion} enables LLM agents to learn from verbal feedback on failed attempts. In reinforcement learning, hindsight experience replay~\cite{her} relabels failed trajectories with achieved goals. Our approach differs: we use failed diagnostic \emph{command sequences} as input to a corrective model that generates improved plans. GRPO~\cite{grpo} provides critic-free optimization suitable for tasks with multiple valid corrections, following the broader trend of RL-based LLM alignment~\cite{instructgpt, ppo, dpo, deepseek_r1}.


%% file: section/03-05-security_Approach.tex
\section{AOI Runtime Architecture}
\label{sec:runtime}

\begin{figure*}[t]
  \centering
  \includegraphics[width=\textwidth]{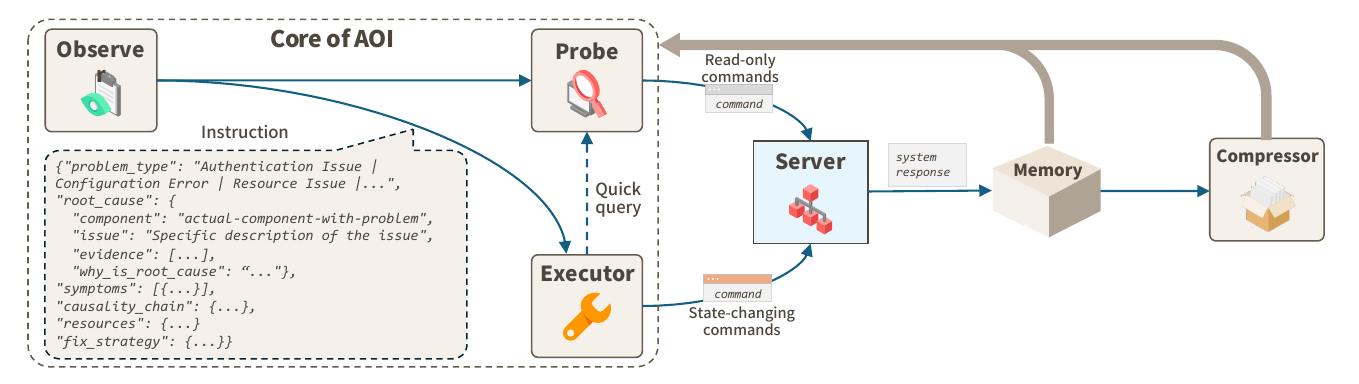}
  \caption{\textbf{AOI Runtime Agent Architecture.} The Observer coordinates diagnosis through the Probe (read-only) and Executor (write-gated) agents. The Compressor maintains context efficiency via dual-timescale memory.}
  \label{fig:aoi}
\end{figure*}

The \method runtime agent system embodies three design principles: (1) \textbf{Safety through Separation}---read-only diagnosis decoupled from state mutation, following the principle of least privilege~\cite{least_privilege}; (2) \textbf{Context Efficiency}---verbose outputs compressed while preserving critical evidence, addressing the well-known context degradation in long inputs~\cite{lost_in_middle}; (3) \textbf{Long-Horizon Coherence}---dual-timescale memory~\cite{memgpt}maintains hypotheses across iterations. Figure~\ref{fig:aoi} illustrates the runtime architecture.

\subsection{Agent Components and Permissions}

\method employs four specialized agents in a structured multi-agent architecture~\cite{multiagent_debate} (Table~\ref{tab:agents}):

\begin{table}[t] 
\centering
\caption{Agent roles and permissions in \method. Right columns show the access control matrix over three memory stores (R = read, W = write, - = no access): $\mathcal{M}_{raw}$ stores raw environment outputs (e.g., full kubectl responses); $\mathcal{M}_{task}$ maintains the diagnostic task queue and hypothesis state; $\mathcal{M}_{comp}$ holds compressed context produced by the Compressor, serving as the Observer's sole information source.}
\label{tab:agents}
\resizebox{\linewidth}{!}{
\begin{tabular}{lll|ccc}
\toprule
\textbf{Agent} & \textbf{Responsibility} & \textbf{Permission} & $\mathcal{M}_{raw}$ & $\mathcal{M}_{task}$ & $\mathcal{M}_{comp}$\\ 
\midrule
Observer ($\mathcal{O}$) & Planning, hypothesis tracking & Read compressed & - & R/W & R \\
Probe ($\mathcal{P}$)  & Diagnostic exploration & Read-only exec & W & R & - \\
Executor ($\mathcal{X}$) & Remediation actions & Write (gated) & W & R & - \\
Compressor ($\mathcal{C}$) & Context distillation & Read raw, write comp. & R & - & W \\
\bottomrule
\end{tabular}
}
\end{table}

\textbf{Observer ($\mathcal{O}$)} serves as the central coordinator. It maintains a diagnostic task queue, analyzes compressed evidence to update hypotheses, and dispatches either Probe or Executor based on current diagnostic state. Critically, the Observer \emph{never directly interacts with the environment}---it only reasons about what to do next.

\textbf{Probe ($\mathcal{P}$)} handles all read-only operations: \texttt{kubectl get}, \texttt{describe}, \texttt{logs}, and similar commands. It supports multi-round exploration within a single iteration (up to $K_{max}$ rounds) and maintains a baseline context cache to avoid redundant queries. Probe implements retry mechanisms for transient failures (network timeouts, API rate limits).

\textbf{Executor ($\mathcal{X}$)} manages state-altering actions. Before executing, it can invoke a single-round Probe for verification (``look before you leap''). All commands pass through a whitelist filter. Executor implements two-stage error recovery: first analyze the failure, then generate a corrected command. All actions are logged for audit.

\textbf{Compressor ($\mathcal{C}$)} bridges raw outputs and decision-making. It applies rule-based deduplication (collapsing repeated log lines) followed by LLM-based semantic compression. Since transformer-based models~\cite{transformer} suffer from degraded attention to middle-context information~\cite{lost_in_middle}, the Compressor uses a sliding window strategy for outputs exceeding context limits and enforces strict token budgets per iteration.


Safety emerges from strict access control between agents and memory stores: Key invariants: (1) Observer cannot read raw outputs---it only sees what Compressor produces, preventing information overload and ensuring consistent inputs; (2) Probe and Executor write to raw store but cannot read it---all information flows through Compressor; (3) Compressor is stateless, processing each iteration independently to avoid error accumulation.


\subsection{Runtime Pipeline}

\subsubsection{Execution Pipeline}
Each iteration follows a four-stage pipeline:
\begin{enumerate}[leftmargin=*,topsep=2pt,itemsep=1pt]
\item \textbf{Decision}: Observer analyzes $(H_{n-2}, C_{n-1})$, produces action type $a_n \in \{\text{Probe}, \text{Execute}, \text{Submit}\}$ and instruction $I_n$.
\item \textbf{Interaction}: Platform routes to appropriate agent; agent executes commands in environment $\mathcal{E}$ and writes raw outputs to $\mathcal{M}_{raw}$.
\item \textbf{Compression}: Compressor processes raw outputs into $C_n$ within token budget.
\item \textbf{Caching}: Platform stores $C_n$ for next iteration; Observer produces summary $S_{n-1}$ appended to $H$.
\end{enumerate}

This pipeline ensures ``raw evidence $\to$ compression $\to$ decision input,'' enforcing budgets at each stage.

\subsubsection{Dual-Timescale Memory}
Long-horizon diagnosis requires maintaining coherence across iterations while respecting context limits. We achieve this through dual-timescale memory:
\textbf{Long-term memory} stores semantic summaries of all past iterations: $H_{n-2} = \{S_1, S_2, \ldots, S_{n-2}\}$. Each summary $S_i$ captures the key findings and hypothesis updates from iteration $i$.
\textbf{Short-term memory} contains the full compressed context from the previous iteration: $C_{n-1} = \text{Compress}(\text{RawOutputs}_{n-1})$.

At iteration $n$, the Observer receives $(H_{n-2}, C_{n-1})$---historical context for continuity plus recent details for informed decisions. This design bounds memory growth while preserving reasoning coherence.

\subsection{Observer Step-Level Policy Optimization}
\label{sec:observer-training}

\method applies GRPO at two distinct granularities to train two different components. We briefly contrast them before detailing the Observer's formulation:

\begin{itemize}[leftmargin=*,topsep=2pt,itemsep=1pt]
\item \textbf{Observer GRPO} (this section) optimizes \emph{step-level decisions}: at each diagnostic iteration, the Observer must choose what action to take next and generate the corresponding context. The reward evaluates whether a single decision effectively advances diagnosis given the current evidence.
\item \textbf{Evolver GRPO} (Section~\ref{sec:evolver}) optimizes \emph{trajectory-level generation}: given a complete seed trajectory, the Evolver produces an improved command sequence covering the entire diagnostic workflow. The reward evaluates the quality of the complete output trajectory.
\end{itemize}

Both share the same GRPO algorithmic framework---sampling $G$ candidates per input and computing group-normalized advantages---but differ in three key aspects: (1) \textbf{optimization granularity} (single decision vs.\ complete trajectory); (2) \textbf{reward design} (six-dimensional step quality vs.\ four-dimensional trajectory quality); and (3) \textbf{training objective} (learning diagnostic reasoning patterns vs.\ learning to augment and repair command sequences). We detail the Observer formulation below; the Evolver formulation follows in Section~\ref{sec:evolver}.

\subsubsection{GRPO Formulation}

For diagnostic tasks, multiple actions can be correct at each step---checking pod logs vs.\ describing deployments may both yield useful information. Unlike standard PPO~\cite{ppo} which requires a learned value function, or DPO~\cite{dpo} which requires pairwise preference data, we optimize the Observer with GRPO~\cite{grpo} using within-group comparisons rather than absolute rewards.

The training data is derived from successful trajectories that are first processed by a Purifier agent (Figure~\ref{fig:overview}), which strips redundant commands (retries, dead-end explorations) and retains only the minimal command sequence leading to correct diagnosis. For each observation context $x$ (comprising compressed evidence and task state), we sample a group of $G$ candidate actions $\{y_i\}_{i=1}^G$ from the Observer policy. Each candidate is scored by an LLM judge $R(x, y_i) \in [0, 1]$. We compute group-normalized advantages:

\begin{equation}
A_i = \frac{R(x, y_i) - \mu_G}{\sigma_G + \epsilon}, \quad \mu_G = \frac{1}{G}\sum_{j=1}^G R(x, y_j)
\end{equation}

The policy gradient update maximizes expected advantage:
\begin{equation}
\nabla_\theta J = \mathbb{E}_{x, \{y_i\}} \left[ \sum_{i=1}^G A_i \nabla_\theta \log \pi_\theta(y_i | x) \right]
\end{equation}

The complete training procedure is detailed in Algorithm~\ref{alg:observer-grpo} (Appendix~\ref{app:observer-grpo}).

\subsubsection{Multi-Dimensional Reward Function}

Unlike the Evolver, which evaluates complete corrected trajectories, the Observer reward operates at the \emph{step level}: at each iteration $n$, an LLM judge~\cite{llm_judge} scores the Observer's decision output across six dimensions. The total reward is a weighted sum of normalized dimension scores:
\begin{equation}
R(x, y) = \sum_{d \in \mathcal{D}} w_d \cdot \frac{s_d}{10}, \quad \sum_{d \in \mathcal{D}} w_d = 1
\end{equation}
where $s_d \in [0, 10]$ is the raw score for dimension $d$ and $w_d$ is its weight. The six dimensions and their default weights are:

\begin{itemize}[leftmargin=*,topsep=2pt,itemsep=1pt]
\item \textbf{Format} ($w{=}0.10$, rule-based): Whether the output is valid JSON. Single clean output scores 10; multiple/repeated JSON blocks score 5; parse failure scores 0 and triggers a hard penalty ($R{=}0.09$).
\item \textbf{Summary} ($w{=}0.15$, LLM): Accuracy of the previous-iteration summary---does it cite specific error messages, component names, and status rather than vague speculation?
\item \textbf{Action} ($w{=}0.10$, LLM): Correctness of the next action type (Probe / Executor / Submit) against the ground-truth action.
\item \textbf{Context Instruction} ($w{=}0.30$, LLM): Quality of the diagnostic reasoning in probe or executor context---does it logically follow from execution history and effectively advance diagnosis?
\item \textbf{Context Namespace} ($w{=}0.30$, LLM): Accuracy of targeted resources (namespaces, pods, services)---are they the right direction for solving the problem at this stage?
\item \textbf{Confidence} ($w{=}0.05$, LLM): Calibration of the self-reported confidence score relative to the iteration stage (e.g., high confidence at iteration 1 is penalized).
\end{itemize}

Context Instruction and Context Namespace together account for 60\% of the total weight, reflecting our emphasis on \emph{diagnostic reasoning quality} and \emph{target accuracy}---the two factors that most directly determine information gain per step.

\section{Trajectory Evolver}
\label{sec:evolver}

While Observer training improves step-level decision-making, a trainable multi-agent system faces a fundamental data scarcity paradox: (1) tasks the system already solves yield trajectories of limited training value---the model gains little from rehearsing what it can already do; (2) tasks the system fails on represent the most valuable improvement targets, yet failed trajectories cannot serve directly as positive examples; (3) high-quality external data (e.g., expert SRE runbooks~\cite{sre_book}) is scarce and expensive to obtain. The Trajectory Evolver resolves this paradox through two mechanisms: \emph{augmenting} successful trajectories into diverse diagnostic workflow variants, and \emph{repairing} failed trajectories into corrected plans that provide directional guidance for future attempts.

\subsection{Problem Formulation}

Given any seed trajectory $\tau = (c_1, r_1, \ldots, c_T, r_T)$, the Evolver learns a policy $\pi_{\text{evolve}}$ that generates an improved command sequence:
\begin{equation}
\tau^{*} = \pi_{\text{evolve}}(\cdot \mid \tau, \text{problem})
\end{equation}

For failed seeds ($\tau^{-}$), this constitutes \emph{repair}: the Evolver corrects diagnostic errors while preserving valid reasoning steps. Failed trajectories provide contrastive supervision---we know what does not work, and often the failures are ``near-misses'' that correctly identify the faulty component but apply incorrect remediation. For success seeds ($\tau^{+}$), this constitutes \emph{augmentation}: the Evolver generates alternative diagnostic workflows that achieve the same goal through different command sequences, expanding training diversity from limited expert data.

\subsection{Seeds: Definition and Data Source}

In the Evolver's context, a \textbf{seed} is a complete fault diagnosis command-sequence trajectory $\tau = (c_1, r_1, \ldots, c_T, r_T)$, spanning from initial exploration to final submission. Seeds are categorized by outcome:
\begin{itemize}[leftmargin=*,topsep=2pt,itemsep=1pt]
\item \textbf{Success seeds}: Trajectories that correctly resolved the incident, containing a validated diagnostic path.
\item \textbf{Failed seeds}: Trajectories that did not resolve the incident, but still contain partial diagnostic value (e.g., correct fault localization with incorrect remediation).
\end{itemize}

The Evolver treats two seed types differently:
\begin{itemize}[leftmargin=*,topsep=2pt,itemsep=1pt]
\item \textbf{Augmentation} (success $\to$ diverse variants): From a single successful trajectory, the Evolver generates multiple command-sequence variants that preserve the core diagnostic logic while varying command choices, exploration order, and supplementary steps. This multiplies limited expert data into diverse training corpus.
\item \textbf{Repair} (failed $\to$ corrected plans): Given full context (problem description + failed trajectory + reference success from similar fault types), the Evolver generates a corrected diagnostic plan. These outputs are not guaranteed to solve the problem directly, but provide the correct \emph{diagnostic reasoning flow}---the right investigation direction, command structure, and remediation strategy---that guides the multi-agent system toward success on retry.
\end{itemize}

In production SRE environments, historical incident records from human operators are a natural seed source---they represent validated diagnostic workflows accumulated over operational history. In our experiments, we use Claude Sonnet 4.5 trajectories on \benchmark as a proxy for such expert records. This choice is deliberate: seeds represent \emph{expert-level diagnostic knowledge}, not the multi-agent system's own outputs. The Evolver's design is agnostic to seed provenance---seeds from human SRE runbooks, frontier models, or the system's own historical successes are interchangeable without any architectural change. We use Sonnet 4.5 solely because it provides sufficient high-quality seeds on the benchmark. For GRPO training, we use only success seeds to ensure training data correctness; for inference, both seed types serve as input (details in Section~\ref{sec:experiments}).

\subsection{GRPO-Optimized Trajectory Correction}

We optimize the Evolver using Group Relative Policy Optimization (GRPO), which is well-suited for tasks where multiple corrected plans may be valid.

For each failed trajectory $\tau^{-}$, sample $G$ candidate corrections $\{\tau_i^{+}\}_{i=1}^G$. Score each correction using a reward model that evaluates:
\begin{itemize}[leftmargin=*,topsep=2pt,itemsep=1pt]
\item \textbf{Validity}: Is the corrected plan executable?
\item \textbf{Completeness}: Does it cover necessary diagnostic steps?
\item \textbf{Correctness}: Are the commands syntactically and semantically correct?
\item \textbf{Effectiveness}: Would this plan lead to successful diagnosis?
\end{itemize}

Compute group-normalized advantages:
\begin{equation}
A_i = \frac{R(\tau^{-}, \tau_i^{+}) - \mu_G}{\sigma_G + \epsilon}, \quad \mu_G = \frac{1}{G}\sum_j R(\tau^{-}, \tau_j^{+})
\end{equation}

GRPO's critic-free design~\cite{grpo, deepseek_r1} avoids overfitting to a single ``correct'' diagnostic path, preserving diversity in valid approaches.

\subsection{Integration with AOI system}

Figure~\ref{fig:evolver} (Appendix) illustrates the Evolver's integration with the AOI runtime.

The corrected command list from the Evolver is provided to the Observer as a structured prompt:

\begin{verbatim}
[Corrected Diagnostic Plan]
Based on analysis of the failed attempt,
the following commands should be executed:
1. kubectl get pods -n {namespace}
2. kubectl describe pod {pod-name}
...
\end{verbatim}

This integration allows the Observer to benefit from learned corrections while retaining flexibility to adapt based on actual system responses. The Evolver provides \emph{guidance}, not rigid constraints.

%% file: section/06-experiments.tex
\section{Experiments}
\label{sec:experiments}

\subsection{Experimental Setup}

We evaluate on \benchmark~\cite{aiopslab}, containing 86 incident scenarios across three microservice applications on live Kubernetes~\cite{k8s} clusters. Tasks span four categories: \textbf{Detection} (32), \textbf{Localization} (28), \textbf{RCA} (13), and \textbf{Mitigation} (13).
We compare against: (1) \textbf{AOL-agent}~\cite{aiopslab}, the benchmark's reference ReAct-style~\cite{react} agent; (2) \textbf{\textsc{STRATUS}}~\cite{stratus}, a multi-agent system with specialized detection/diagnosis/mitigation modules; and (3) \textbf{Claude Sonnet 4.5}~\cite{claude} with AOL-agent, whose trajectories serve as seed data for our training pipeline.

\subsubsection{Data Split}

We adopt a nested partition structure illustrated in Figure~\ref{fig:data-split}(Appendix). All three sets share the same execution environment, ensuring consistency across training and evaluation.

\begin{itemize}[leftmargin=*,topsep=2pt,itemsep=1pt]
    \item $\mathcal{D}^{all}$: the full benchmark. Runtime comparison is evaluated over all 86 tasks using the base model \emph{without any task-specific training}, ensuring a fair architectural comparison against STRATUS. The training subsets below are drawn from $\mathcal{D}^{all}$.
    \item $\mathcal{D}^{train}_{evolver}$: 49 tasks successfully solved by Claude Sonnet 4.5, serving as success seeds for Evolver GRPO training.
    \item $\mathcal{D}^{train}_{obs}$: a strict subset of $\mathcal{D}^{train}_{evolver}$ covering 11 fault types and 23 entries, used for Observer GRPO training.
    \item $\mathcal{D}^{test}_{obs}$: 63 held-out evaluation examples for the Observer, covering 15 unseen fault types plus tasks from training fault types where Sonnet failed.
    \item $\mathcal{D}^{test}_{evolver}$: 37 failed trajectories from Claude Sonnet 4.5, a subset of $\mathcal{D}^{test}_{obs}$, used to evaluate the Evolver's repair capability.
\end{itemize}

In the combined pipeline, the Observer receives diagnostic guidance generated by the Evolver. To evaluate this end-to-end effect without data leakage, \emph{every test case must be unseen by both components}. By making the Observer's training set a strict subset of the Evolver's, we guarantee that $\mathcal{D}^{test}_{evolver} \subset \mathcal{D}^{test}_{obs}$---all 37 Evolver test cases fall within the Observer's 63 held-out tasks. This nested design enables fair evaluation of each component independently and in combination.

\textbf{Strict fault-type split for $\mathcal{D}^{train}_{obs}$.} We adopt the most stringent split strategy: training and test fault types have \textbf{zero overlap}. The 86 tasks span 26 distinct fault types; we assign 11 types (38 total tasks, of which 23 have successful Sonnet trajectories) to training and reserve the remaining 15 types exclusively for testing. This ensures that the Observer is evaluated on genuinely novel fault categories, not merely unseen instances of familiar faults. The complete fault-type partition is provided in Appendix~\ref{app:data-split}.


\subsubsection{Metrics.} 
We report success using two standard multi-run aggregation metrics:
\textbf{best@$k$}, where a task is counted as solved if any of $k$ independent runs succeeds; and
\textbf{avg@$k$}, the mean success rate across $k$ runs.

\subsubsection{Implementation.} 
Our method is built on Qwen3-14B~\cite{qwen3} as the base model. We apply LoRA~\cite{lora} fine-tuning with rank 64, $\alpha$=128, and learning rate $10^{-5}$; GRPO~\cite{grpo} is configured with group size $G$=4. All experiments run on 2$\times$A100 GPUs with vLLM~\cite{vllm} for inference.

\subsection{\method Runtime vs Baseline}

We compare \method against STRATUS on $\mathcal{D}^{all}$ (86 tasks) with identical base models. No task-specific training.

\begin{table*}[t]
\centering
\caption{\method runtime comparison on full benchmark (86 tasks, \%). GPT-4o-mini~\cite{gpt4}, Claude Sonnet 4.5 uses single-run; Qwen3-14B uses 5-round sampling (best@5/avg@5).}
\label{tab:runtime}
\small
\begin{tabular}{llccccc}
\toprule
\textbf{Model} & \textbf{Method} & \textbf{Detection} & \textbf{Localization} & \textbf{RCA} & \textbf{Mitigation} & \textbf{Overall} \\
\midrule
GPT-4o-mini & AOL-agent & 25.0 & 9.5 & 7.7 & 7.7 & 14.7 \\
GPT-4o-mini & STRATUS & 78.1 & 25.0 & 15.4 & 23.1 & 43.0 \\
GPT-4o-mini & \method & \textbf{90.6} & \textbf{32.1} & \textbf{38.5} & \textbf{53.8} & \textbf{58.1} \\
\midrule
Claude Sonnet 4.5 & AOL-agent & 68.8 & 53.6 & 15.4 & 76.9 & 57.0 \\
\midrule
Qwen3-14B (best@5/avg@5) & STRATUS & 75.0/41.3 & 32.1/11.4 & 7.7/4.6 & 15.4/15.4 & 41.9/22.1 \\
Qwen3-14B (best@5/avg@5) & \method & \textbf{100/66.9} & \textbf{53.6/27.9} & \textbf{30.8/7.7} & \textbf{46.2/23.1} & \textbf{66.3/38.6} \\
\bottomrule
\end{tabular}
\end{table*}

\textbf{Architecture yields 4$\times$ improvement over vanilla agents.} \method with GPT-4o-mini achieves 58.1\% overall compared to AOL-agent's 14.7\%. The gains stem from read-write separation: the Observer can safely explore diagnostic paths without risking state mutations, while the Executor applies changes only after sufficient evidence accumulates.

\textbf{Task complexity reveals architectural bottlenecks.} Detection benefits most from safe exploration (100\% best@5 vs STRATUS's 75\%), as agents can issue multiple diagnostic commands without penalty. RCA shows the largest relative gains (+150\% over STRATUS), indicating that \method's dual-timescale memory helps maintain reasoning coherence across long diagnostic chains. Mitigation improves 3$\times$ over STRATUS because Executor-level safety gates prevent the cascading failures that occur when agents attempt remediation before completing diagnosis.

\textbf{Frontier models excel at pattern-matching, not reasoning.} Claude Sonnet 4.5 achieves 76.9\% on Mitigation---the highest single-category result---but only 15.4\% on RCA, matching GPT-4o-mini + STRATUS. This asymmetry reveals that Sonnet's strength lies in recognizing remediation patterns (restart pods, scale replicas, rollback deployments) rather than multi-step causal reasoning. Mitigation tasks have well-defined action templates once the fault is identified; RCA requires synthesizing signals across logs, metrics, and traces to establish causality. \method's structured diagnostic workflow addresses this gap: with Qwen3-14B, RCA doubles to 30.8\% while Mitigation reaches 46.2\%.

\textbf{Open-weight surpasses frontier with proper architecture.} Qwen3-14B + \method (66.3\% best@5) outperforms Claude Sonnet 4.5 + AOL-agent (57.0\%), solving 57 vs 49 tasks. The 14B model's disadvantage in raw capability is offset by architectural support for systematic exploration and safe execution. Diminishing returns analysis (Figure~\ref{fig:bestk}, Appendix) shows best@1$\to$2 gains +19.8 points while best@3$\to$5 adds only +8.2 combined, suggesting 2-3 sampling rounds capture most benefit.






\subsection{Observer GRPO Generalization}

We evaluate whether Observer GRPO training improves generalization to unseen fault types.

\subsubsection{Held-out Fault Types}

We train Observer on $\mathcal{D}^{train}_{obs}$ (23 tasks) and evaluate on 63 held-out tasks (unseen fault types).

\begin{table}[t]
\centering
\caption{Observer GRPO on held-out fault types (63 tasks, \%).}
\label{tab:observer}
\small
\begin{tabular}{lccccc}
\toprule
\textbf{Method} & \textbf{Det.$\uparrow$} & \textbf{Loc.$\uparrow$} & \textbf{RCA$\uparrow$} & \textbf{Mit.$\uparrow$} & \textbf{Overall$\uparrow$} \\
 & (22) & (22) & (12) & (7) & (63) \\
\midrule
Sonnet 4.5 (AOL-agent) & 54.5 & 40.9 & 8.3 & 57.1 & 41.3 \\
\midrule
\method (Untrained) & 65.5 & 22.7 & 6.7 & 14.3 & 33.7 \\
\method (Observer-GRPO) & \textbf{90.9} & 18.2 & \textbf{16.7} & 14.3 & \textbf{42.9} \\
\bottomrule
\end{tabular}
\end{table}

\textbf{GRPO learns task-completion strategies, not intermediate accuracy.} Observer-GRPO achieves 90.9\% Detection (+36 points over Sonnet) but Localization drops from 22.7\% to 18.2\%. This trade-off is inherent to GRPO's reward structure: the model optimizes for end-task success, learning that Detection can succeed with coarse-grained anomaly signals while Localization requires precise component identification. The trained Observer prioritizes high-confidence fault indicators over exhaustive exploration. Appendix~\ref{app:rl-analysis} provides comprehensive task-type analysis revealing that GRPO-trained models use $\sim$9 more exploration steps---beneficial for RCA but harmful for Localization where over-exploration causes multi-anomaly confusion.

\textbf{Single-run open-source beats frontier model.} Observer-GRPO (42.9\% avg@1) surpasses Claude Sonnet 4.5 (41.3\%) on identical held-out tasks without multi-run sampling. The improvement concentrates in Detection (+36.4 points) and RCA (+8.4 points)---precisely the tasks requiring systematic diagnostic reasoning rather than pattern recognition. Mitigation remains unchanged at 14.3\%: in \method's architecture, remediation commands are generated and executed by the Executor, so Observer optimization cannot directly improve the quality of the final repair actions. Sonnet's Mitigation advantage (57.1\%) stems from its stronger base capability in generating correct remediation commands.

\textbf{Generalization to unseen fault types validates transfer learning.} The held-out set contains 15 fault types never seen during training. Observer-GRPO's gains demonstrate that diagnostic patterns transfer across fault categories: the model learns \emph{how} to diagnose (command sequencing, signal prioritization) rather than memorizing fault-specific solutions. Per-task analysis (Figure~\ref{fig:grpo}, Appendix) shows consistent improvement across fault types rather than selective memorization.




\subsubsection{Combined System on Failed Cases}

We evaluate all component combinations on $\mathcal{D}^{test}_{evolver}$ (37 Sonnet-failed tasks, 5 independent runs per task): Base (untrained Observer), Evolver-prompts (Base + Evolver-generated diagnostic prompts), Observer-GRPO (trained Observer without Evolver), and Observer-GRPO + Evolver-prompts (trained Observer with Evolver-generated prompts).

\begin{table}[t]
\centering
\caption{Component ablation on $\mathcal{D}^{test}_{evolver}$ (37 tasks, best / avg \%). First three rows use 5 runs; last row uses 4 runs.}
\label{tab:selfevolve}
\small
\begin{tabular}{lccccc}
\toprule
\textbf{Method} & \textbf{Det.} & \textbf{Loc.} & \textbf{RCA} & \textbf{Mit.} & \textbf{Overall} \\
 & (10) & (13) & (11) & (3) & (37) \\
\midrule
Base & 100/50.0 & \textbf{54}/26.2 & 27/7.3 & 0/0 & \textbf{54}/24.9 \\
Evolver-prompts & 90/64.0 & \textbf{54}/24.6 & 18/12.7 & 0/0 & 49/29.7 \\
Observer-GRPO & 90/64.0 & 38/21.5 & \textbf{36}/\textbf{29.1} & 0/0 & 49/33.5 \\
\raisebox{-0.5\normalbaselineskip}[0pt][0pt]{\shortstack[l]{Observer-GRPO\\[-3pt]+ Evolver-prompts}} & \textbf{100}/72.5 & 31/19.2 & \textbf{36}/25.0 & 0/0 & 49/\textbf{33.8} \\[6pt]
\bottomrule
\end{tabular}
\end{table}

\textbf{Observer-GRPO + Evolver-prompts achieves highest avg} (33.8\%, Table~\ref{tab:selfevolve}), a +8.9 point improvement over Base. The combination is complementary: Observer-GRPO improves RCA (+21.8 points over Base), while Evolver-prompts improves Detection consistency (avg: 50\%$\to$72.5\%).

\textbf{Components target different failure modes.} Observer-GRPO benefits tasks requiring deep exploration (RCA), while Evolver-prompts benefits tasks requiring structured workflows (Detection). Localization shows minimal gains from either---these tasks require precise fault identification that neither exploration depth nor workflow guidance addresses.

\textbf{Localization hits a structural ceiling.} Best@5 Localization is identical across conditions, and avg@5 barely changes. Localization requires pinpointing the exact faulty component, which depends on real-time environment state that offline-generated plans cannot anticipate. The Evolver improves \emph{what} to investigate but not \emph{where} to look.

\textbf{Dual-timescale synergy.} Observer-GRPO + Evolver-prompts combines both components addressing complementary failure modes: Observer-GRPO improves diagnostic \emph{execution} (fast timescale), while Evolver-prompts improves diagnostic \emph{planning} (slow timescale). Their combination recovers nearly a third of cases where Claude Sonnet 4.5 achieved 0\%.

\textbf{Mitigation remains at 0\%.} Unlike Detection and RCA, where structural diagnostic patterns generalize across fault types, Mitigation requires fault-specific remediation actions absent from the training distribution.


\subsection{Repair Quality of Evolver}

\subsubsection{Evaluation Strategies}
We evaluate the Evolver's ability to repair previously failed cases. We train on 49 successful trajectories and test on 37 cases where Claude Sonnet originally failed.The complete command sequences produced by the Evolver are ``adapted'' versions of the original trajectories---the repaired command lists may differ from the original AIOpsLab problem context (e.g., adjusted command arguments, alternative exploration paths), making direct replay in the environment infeasible. We therefore adopt two complementary evaluation strategies:

\begin{enumerate}[leftmargin=*,topsep=2pt,itemsep=1pt]
\item \textbf{Evolver-as-Prompt (end-to-end validation):} We feed the Evolver's corrected command sequences as prompts to the Observer and measure whether actual task success improves (see Table~\ref{tab:selfevolve}, overall increasing 4.8\% on avg@5).
\item \textbf{LLM judge scoring (generation quality evaluation):} We use Claude Opus 4.5 to score the Evolver's repaired outputs across four dimensions---Validity, Completeness, Correctness, and Effectiveness---to assess repair quality improvement.
\end{enumerate}

\subsubsection{Comparison Experiments}
\begin{figure*}[t]
  \centering
  \begin{minipage}[t]{0.48\textwidth}
    \centering
    \includegraphics[width=\linewidth]{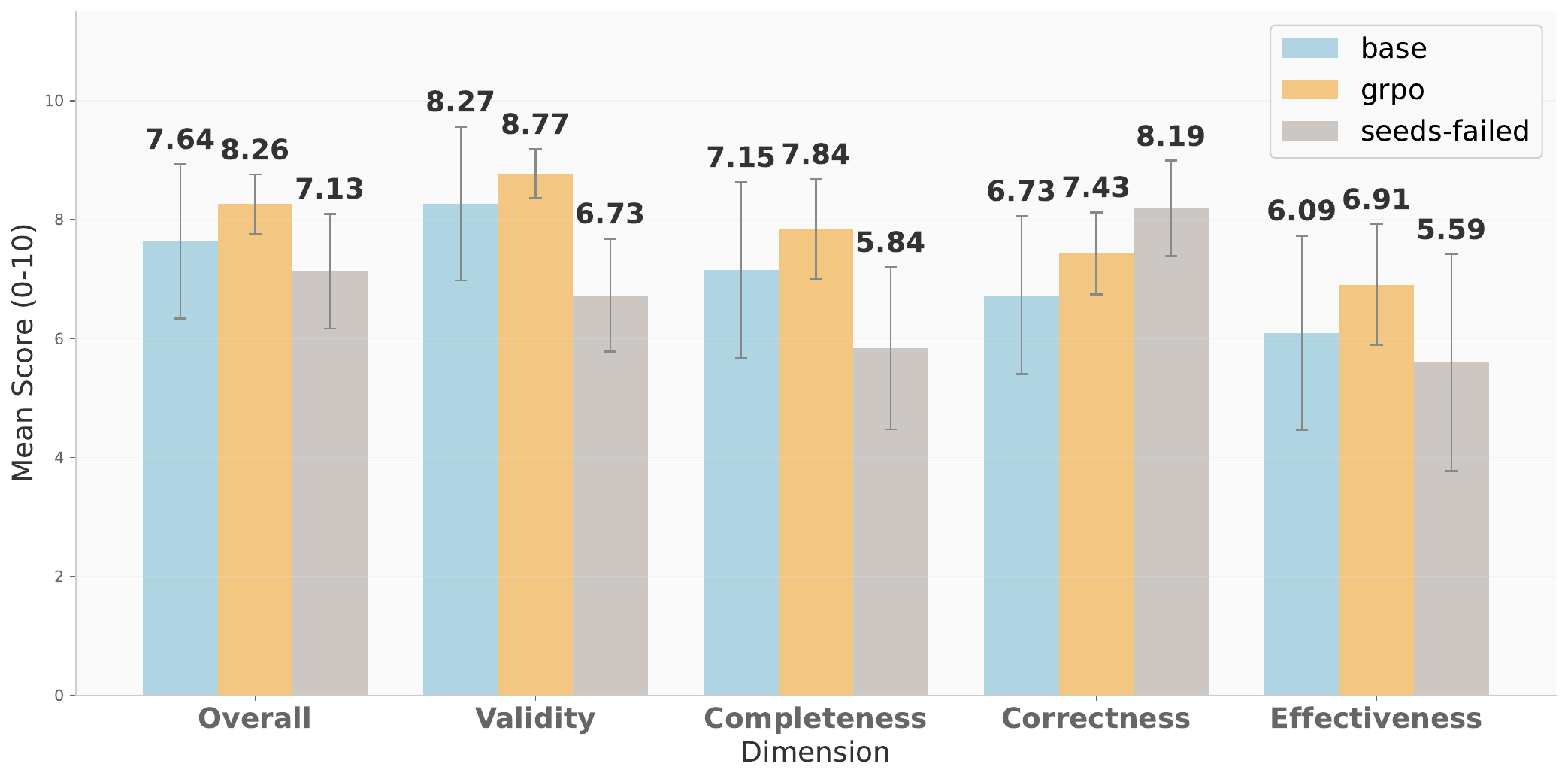}
    \caption{\textbf{Performance across evaluation dimensions.} Seeds-Failed, Base (untrained Qwen3-14B), and GRPO-trained Evolver compared on four reward dimensions and overall score.}
    \label{fig:main-results}
  \end{minipage}
  \hfill
  \begin{minipage}[t]{0.48\textwidth}
    \centering
    \includegraphics[width=\linewidth]{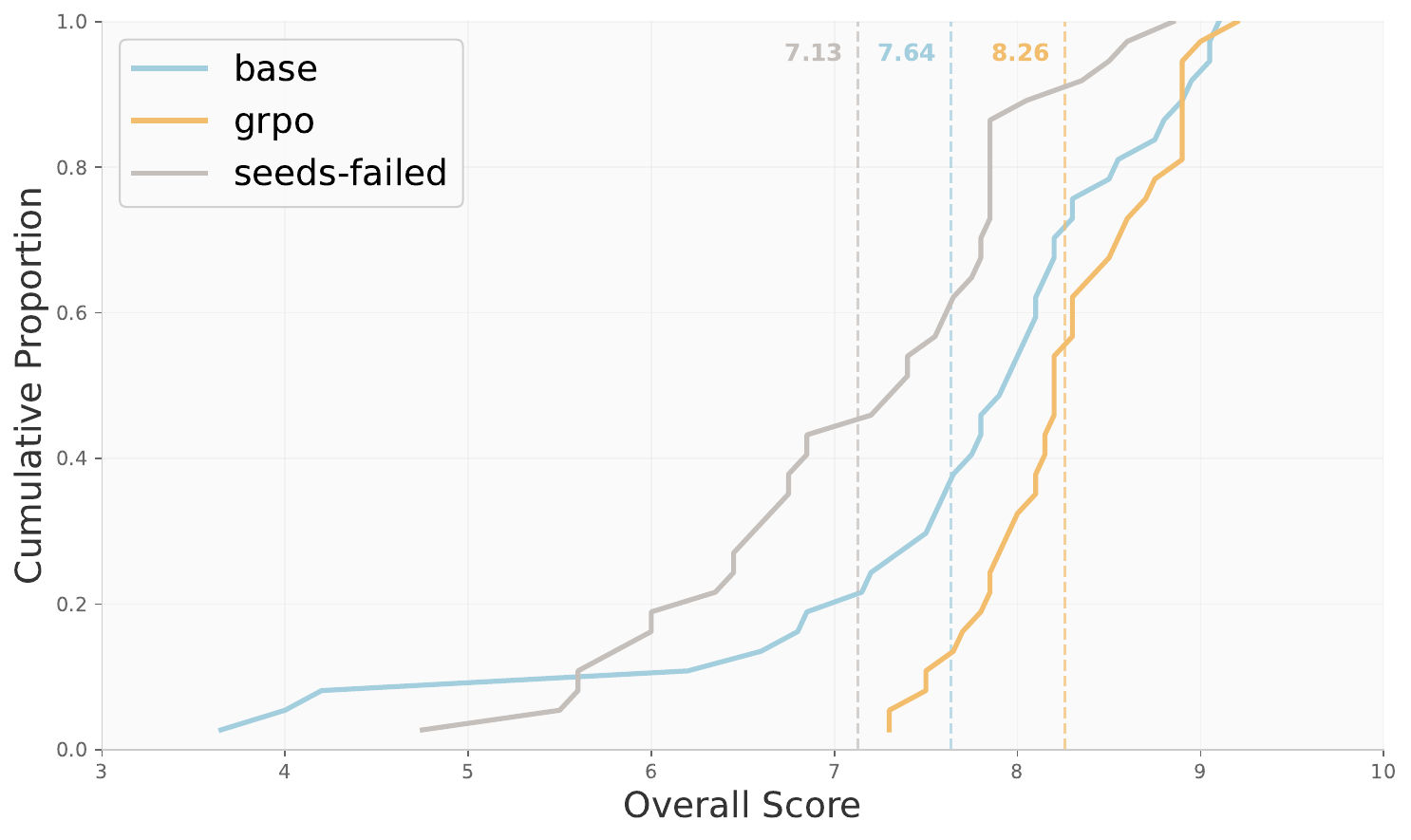}
    \caption{\textbf{Cumulative distribution of overall scores.} Dashed lines indicate group means.}
    \label{fig:distribution}
  \end{minipage}
\end{figure*}

Three groups of experiment are compared in Figure~\ref{fig:main-results} and Figure~\ref{fig:distribution} are:
\begin{itemize}[leftmargin=*,topsep=2pt,itemsep=1pt]
\item \textbf{Seeds-Failed}: The original command sequences from Sonnet 4.5's 37 failed interactions with AIOpsLab.
\item \textbf{Base}: The untrained Qwen3-14B model repairing each of the 37 tasks, generating 2 repair candidates per task.
\item \textbf{GRPO}: The GRPO-trained Qwen3-14B Evolver repairing each of the 37 tasks using the same Seeds-Failed as prompt input, generating 2 repair candidates per task.
\end{itemize}

\subsubsection{Quality Evaluation Results}
In Figure~\ref{fig:main-results}, although Seeds-Failed retains higher correctness due to pre-execution filtering, the Evolver models accept minor syntactic variations to achieve diverse diagnostic paths, yielding net gains in Completeness and Effectiveness that outweigh the slight correctness decline. Crucially, improvements in Validity and Completeness indicate that GRPO facilitates structural rather than superficial correction, teaching the Evolver to fill logical gaps in diagnostic reasoning instead of memorizing command strings . This stability is evidenced by the Cumulative Distribution Function (CDF) in Figure~\ref{fig:distribution}, where the elimination of the lower-bound tail and a variance reduction (std $0.97 \to 0.49$) confirm the acquisition of consistent strategies rather than stochastic improvements . By building upon the Base model's non-trivial floor to amplify repair consistency, GRPO proves superior to direct supervised fine-tuning in accommodating diverse valid diagnostic paths .

\subsubsection{End-to-End Robustness Analysis (Table~\ref{tab:selfevolve})}

\textbf{Evolver-prompts improves robustness, not peak performance.} The most revealing pattern in Table~\ref{tab:selfevolve} lies in the gap between best@5 and avg@5---a direct measure of run-to-run variance. For Base, this gap is 29.2pp (54.1\%$-$24.9\%); for Evolver-prompts, it shrinks to 18.9pp (48.6\%$-$29.7\%). While best@5 decreases slightly ($-$5.5pp), avg@5 improves meaningfully (+4.8pp). Evolver-prompts' structural prompts act as \emph{diagnostic scaffolding} that constrains the Observer's search space, making successful diagnostic paths reliably reproducible.

\textbf{Detection: variance halved.} Without Evolver prompts, the Observer achieves 100\% best@5 but only 50\% avg@5---a 50pp gap indicating that half of all runs fail despite the problem being solvable. With Evolver prompts, best@5 drops to 90\% but avg@5 jumps to 64\%, compressing the gap to 26pp.

\textbf{Analysis benefits disproportionately from structural guidance.} Although Analysis best@5 decreases (27.3\%$\to$18.2\%), avg@5 nearly doubles (7.3\%$\to$12.7\%). This divergence is informative: Analysis tasks---root cause identification requiring multi-step reasoning chains---are the most sensitive to diagnostic plan quality. Without prompts, the Observer occasionally stumbles upon correct causal chains (high best@5) but cannot reliably reproduce them (low avg@5). The Evolver's prompts provide a reasoning scaffold that makes successful diagnosis \emph{repeatable} rather than accidental, which is precisely the property needed for production deployment where consistent reliability matters more than occasional brilliance.

%% file: section/07-discussion.tex
\section{Discussion}
\label{sec:discussion}

\textbf{Failed trajectories encode recoverable signal.} The conventional wisdom treats failed diagnostic attempts as noise to be filtered. Our experiments suggest otherwise: the GRPO-trained Evolver significantly improves repair quality on previously failed cases (mean LLM judge score 7.18$\to$8.27, std 0.97$\to$0.49) by learning \emph{what was almost right}. Failed trajectories often contain correct diagnostic intuitions paired with execution errors---wrong command flags, incorrect resource names, or missing intermediate steps. GRPO's contrastive learning surfaces these near-miss patterns, enabling systematic correction rather than wholesale replacement.

\textbf{Diminishing returns in sampling diversity.} Our best@$k$ analysis reveals a surprising pattern: success rate improves by 19.8 points from one to two runs, but only 8.2 points across the subsequent three runs combined (Figure~\ref{fig:bestk}). This suggests that diagnostic stochasticity is bounded---most tasks either succeed consistently, fail consistently, or exhibit binary variance that two attempts suffice to capture. For practitioners, this implies that allocating compute budget to 2-3 independent runs yields better cost-effectiveness than deeper single-run optimization.

\textbf{The speed-precision trade-off in learned diagnostics.} Observer GRPO improves Detection by 25.5 points while \emph{degrading} Localization by 4.5 points. Task-level analysis reveals the mechanism: GRPO learns aggressive diagnostic shortcuts that terminate earlier with correct anomaly verdicts, but occasionally skip the detailed exploration steps required for precise fault localization. The two degraded tasks (pod\_failure\_hotel\_res, astronomy\_shop\_product\_catalog) both require multi-hop reasoning through service dependencies---exactly the exploration that fast paths eliminate. This trade-off is not a bug but a feature of reward-driven optimization: the training signal emphasized detection accuracy, and the model responded accordingly.

\textbf{Safety mechanisms improve capability.} Counter-intuitively, constraining the agent's action space \emph{enhances} diagnostic success. Read-write separation forces evidence accumulation before mutation, preventing the cascading failures we observed in STRATUS where premature remediation attempts corrupted system state. This challenges the assumption that safety constraints reduce capability---for operational tasks with irreversible actions in production environments~\cite{k8s, sre_book}, guardrails enable more aggressive exploration within safe boundaries.

\textbf{Capability boundaries are task-specific.} One-third of tasks (29/86) fail consistently across all configurations---five Qwen rounds, GPT-4o-mini, and GRPO variants. These are not random failures but systematic capability gaps: MongoDB authentication recovery requires Helm-specific knowledge absent from training, and multi-service localization demands causal reasoning beyond current architectures. Importantly, these boundaries are stable and predictable, enabling practitioners to identify which incident types require human escalation.

%% file: section/08-limitations.tex
\section{Limitations}
\label{sec:limitations}



\noindent \textbf{Expanding the Evolver.} While the current Evolver generates corrected command sequences as structured prompts for LLMs, its role could be extended to produce synthetic system feedback via environment simulators or to serve as a runtime agent for dynamic plan refinement. Such architectural expansions, while promising for further boosting system autonomy, are beyond the scope of this work and are left for future investigation.

\noindent \textbf{Beyond AIOpsLab.} Although \benchmark offers a realistic evaluation of SRE scenarios, we anticipate the emergence of broader community benchmarks covering more diverse infrastructure stacks and larger-scale deployments. On the applied side, we plan to deploy \method in production SRE environments to validate the use and productization potential of the framework within real-world incident response workflows.

%% file: section/09-conclusion.tex
\section{Conclusion}
\label{sec:conclusion}

We presented \method, a system for autonomous cloud incident response built on two innovations: (1) an Observer-Probe-Executor runtime that enforces safety through architectural separation, and (2) a Trajectory-Corrective Evolver that learns to correct failed diagnostic sequences via GRPO optimization.
Our results demonstrate a key insight: \emph{failed trajectories are not wasted supervision}. By learning to correct failures rather than discard them, we convert 37 failed cases into training signal that enables systematic capability acquisition. The Evolver improves end-to-end avg@5 by 4.8\% while reducing run-to-run variance by 35\% (best@5--avg@5 gap: 29.2pp$\to$18.9pp), and the GRPO-trained Evolver achieves higher repair quality (mean LLM judge score 7.18$\to$8.27, std 0.97$\to$0.49), indicating robust learning of correction patterns.

%% file: section/10-appendix-revised-feb8.tex
\section*{Code and Data Availability}
  \label{app:code}
  Code and pre-trained models: \url{https://anonymous.4open.science/r/aoi-C8C7}

\textbf{Benchmark data.} We use the publicly available AIOpsLab benchmark~\cite{aiopslab}. Our fault-type split and trajectory data will be released upon publication.
   
\noindent\textit{Note: Anonymous GitHub repository link will be provided for review. The repository will be made public upon paper acceptance.}
\section{Algorithm Details}
\label{app:algorithms}

\subsection{AOI Main Loop}

Algorithm~\ref{alg:main} presents the complete \method orchestration loop. The design reflects three key principles:

\textbf{Dual-timescale memory.} The algorithm maintains two memory structures: (1) \emph{short-term context} $C_n$ from the Compressor, preserving critical evidence from the current iteration, and (2) \emph{long-term memory} $H$ storing summaries $S_{n-1}$ across iterations. This separation allows the Observer to maintain hypotheses over long diagnostic horizons while keeping per-iteration context within token limits.

\textbf{Lazy compression.} Compression occurs at the end of each iteration (line 17) rather than before decision-making. This ensures the Observer sees raw evidence when making decisions, while only compressed summaries are stored for future iterations---balancing information fidelity with context efficiency.

\textbf{Action-type routing.} The three action types (Submit, Probe, Execute) route to specialized agents with different capabilities and safety constraints, enabling fine-grained access control.

\begin{algorithm}[t]
\caption{\method Main Loop}
\label{alg:main}
\small
\begin{algorithmic}[1]
\STATE \textbf{Input:} Task $\mathcal{D}$, max iterations $N$
\STATE Initialize Observer $\mathcal{O}$, Probe $\mathcal{P}$, Executor $\mathcal{X}$, Compressor $\mathcal{C}$
\STATE $\mathcal{T} \gets \mathcal{O}.\text{Plan}(\mathcal{D})$ \hfill $\triangleright$ Initial task queue
\STATE $H \gets []$ \hfill $\triangleright$ Long-term memory
\FOR{$n = 1$ to $N$}
    \STATE $(S_{n-1}, a_n, I_n) \gets \mathcal{O}.\text{Decide}(H_{n-2}, C_{n-1}, \mathcal{T})$
    \IF{$n > 1$}
        \STATE $H.\text{append}(S_{n-1})$
    \ENDIF
    \IF{$a_n = \text{Submit}$}
        \STATE \textbf{return} $\mathcal{E}.\text{Submit}(I_n)$
    \ELSIF{$a_n = \text{Probe}$}
        \STATE $\mathcal{P}.\text{Run}(I_n, \mathcal{E}) \to \mathcal{M}_{raw}$
    \ELSIF{$a_n = \text{Execute}$}
        \STATE $\mathcal{X}.\text{Run}(I_n, \mathcal{E}) \to \mathcal{M}_{raw}$
    \ENDIF
    \STATE $C_n \gets \mathcal{C}.\text{Compress}(\mathcal{M}_{raw}^{(n)})$
\ENDFOR
\STATE \textbf{return} $\mathcal{E}.\text{Submit}(\text{timeout})$
\end{algorithmic}
\end{algorithm}

\subsection{Observer GRPO Training}
\label{app:observer-grpo}

Algorithm~\ref{alg:observer-grpo} presents the Observer GRPO training procedure.

\begin{algorithm}[htbp]
\caption{Observer GRPO Training}
\label{alg:observer-grpo}
\small
\begin{algorithmic}[1]
\STATE \textbf{Input:} Successful trajectories $\mathcal{D}$, group size $G$, LLM judge $R$
\STATE \textbf{Output:} Trained Observer policy $\pi_\theta$
\FOR{each training step}
    \STATE Sample context $x$ from $\mathcal{D}$ (compressed evidence + task state)
    \STATE Sample $G$ candidate actions $\{y_i\}_{i=1}^G \sim \pi_\theta(\cdot | x)$
    \FOR{$i = 1$ to $G$}
        \STATE $r_i \gets R(x, y_i)$ \hfill $\triangleright$ LLM judge scoring
    \ENDFOR
    \STATE $\mu_G \gets \frac{1}{G}\sum_i r_i$; \quad $\sigma_G \gets \sqrt{\frac{1}{G}\sum_i (r_i - \mu_G)^2}$
    \STATE $A_i \gets (r_i - \mu_G) / (\sigma_G + \epsilon)$ for all $i$
    \STATE Update $\theta$ via policy gradient with advantages $\{A_i\}$
\ENDFOR
\end{algorithmic}
\end{algorithm}

The LLM judge evaluates candidates on six dimensions (consistent with Section~\ref{sec:observer-training}):
\begin{itemize}[leftmargin=*,topsep=2pt,itemsep=1pt]
\item \textbf{Format} ($w{=}0.10$, rule-based): Whether the output is valid JSON.
\item \textbf{Summary} ($w{=}0.15$, LLM): Accuracy of the previous-iteration summary.
\item \textbf{Action} ($w{=}0.10$, LLM): Correctness of the next action type (Probe / Executor / Submit).
\item \textbf{Context Instruction} ($w{=}0.30$, LLM): Quality of the diagnostic reasoning in probe or executor context.
\item \textbf{Context Namespace} ($w{=}0.30$, LLM): Accuracy of targeted resources (namespaces, pods, services).
\item \textbf{Confidence} ($w{=}0.05$, LLM): Calibration of self-reported confidence relative to iteration stage.
\end{itemize}

\textbf{Rationale for dimension weights.} Context Instruction and Context Namespace together account for 60\% of the total weight, reflecting the emphasis on diagnostic reasoning quality and target accuracy---the two factors that most directly determine information gain per step. Format is rule-based to provide a hard constraint on output validity. Summary, Action, and Confidence serve as auxiliary signals that shape the learning trajectory without dominating the reward.

\subsection{Trajectory-Corrective Evolver}

The Evolver addresses a key limitation of pure imitation learning: when the Observer encounters situations not covered by successful
  trajectories, it lacks guidance for recovery. Given a failed trajectory, the Evolver generates candidate corrections via GRPO sampling and
  selects the highest-scoring correction as a structured prompt for the Observer's next attempt. This creates a closed-loop mechanism that
  converts failed diagnostic attempts into learning opportunities without requiring manual expert intervention.

\begin{figure}[htbp]
  \centering
  \includegraphics[width=\columnwidth]{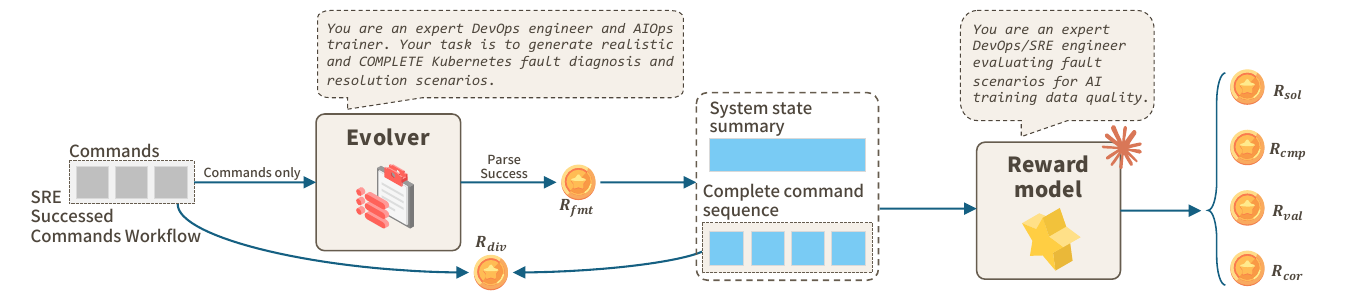}
  \caption{\textbf{Trajectory-Corrective Evolver Architecture.} The Evolver observes failed command sequences from the Observer's execution history, generates $G$ candidate corrections via GRPO sampling (sampling from a policy trained on successful trajectories), and provides the highest-scoring correction as a structured prompt to guide the Observer's next attempt. \emph{Key insight:} This closed-loop mechanism converts failed diagnostic attempts into learning opportunities without requiring manual expert intervention.}
  \label{fig:evolver}
\end{figure}

  Figure~\ref{fig:evolver} illustrates the Evolver's integration with the AOI runtime.

\noindent\textit{Figure explanation:} The Evolver operates in three stages: (1) \textbf{Failure Collection} - collects failed trajectories
  from previous diagnostic attempts; (2) \textbf{Correction Generation} - samples multiple candidate corrected command sequences using
  GRPO-trained policy; (3) \textbf{Guidance Injection} - provides the highest-scoring corrected plan as a structured prompt to the Observer for
   subsequent attempts. The dashed feedback loop shows how this process converts failed trajectories into training signal.

\section{Experimental Setup}
\label{app:setup}

\subsection{Hyperparameters}
\label{app:hyperparams}

\textbf{Key parameter choices.} We set max iterations to 15 based on analysis of successful Claude Sonnet trajectories, where 95\% completed within 12 iterations. The context budget (4096 tokens) balances information retention with inference cost---larger budgets showed diminishing returns beyond 4K tokens. For GRPO training, we use LoRA rank 64 (rather than full fine-tuning) to preserve the base model's general capabilities while adapting to diagnostic patterns. The learning rate ($10^{-5}$) and batch size (16) were selected to ensure stable convergence within 3 epochs.

\begin{table}[htbp]
\centering
\caption{Complete hyperparameter settings.}
\small
\begin{tabular}{ll}
\toprule
\textbf{Parameter} & \textbf{Value} \\
\midrule
\multicolumn{2}{l}{\emph{AOI Runtime}} \\
Max iterations & 15 \\
Max Probe rounds/iteration & 5 \\
Context budget/iteration & 4096 tokens \\
Long-term memory capacity & 10 summaries \\
Executor whitelist & 47 command patterns \\
\midrule
\multicolumn{2}{l}{\emph{Trajectory-Corrective Evolver}} \\
Base model & Qwen3-14B \\
GRPO candidates $G$ & 4 \\
Learning rate & $1 \times 10^{-5}$ \\
LoRA rank / alpha & 64 / 128 \\
Batch size & 16 \\
Epochs & 3 \\
Reward model & Claude Opus 4.5 \\
Training samples 49 success seeds \\
Test samples & 37 failed seeds \\  
\bottomrule
\end{tabular}
\end{table}

\subsection{Benchmark Details}
\label{app:benchmark}

\textbf{Full benchmark} from \benchmark~\cite{aiopslab}: 88 scenarios across 3 applications (reduced to 86 in our evaluation due to 2 deprecated scenarios).

\textbf{Our evaluation split:}
\begin{itemize}[leftmargin=*,topsep=2pt,itemsep=1pt]
\item \textbf{Success seeds} (49 cases): Claude Sonnet successful trajectories, used as positive examples for GRPO training.
\item \textbf{Failed seeds} (37 cases): Claude Sonnet failed trajectories that overlap with Observer's held-out set, used as test set to evaluate Evolver's recovery ability.
\end{itemize}

Failed cases by task type:
\begin{table}[htbp]
\centering
\small
\begin{tabular}{lc}
\toprule
\textbf{Task Type} & \textbf{Count} \\
\midrule
Detection & 10 \\
Localization & 13 \\
Mitigation & 3 \\
RCA & 11 \\ 
\midrule
Total & 37 \\
\bottomrule
\end{tabular}
\end{table}

Fault types in failed cases include: service failures (31\%), misconfigurations (26\%), authentication errors (18\%), pod failures (15\%), and network issues (10\%).

\subsection{Data Split Details}
\label{app:data-split}

\textbf{Rationale for nested data partition.} We use a strict subset structure $\mathcal{D}^{train}_{obs} \subset \mathcal{D}^{train}_{evolver} \subset \mathcal{D}^{all}$ to ensure no data leakage: since the Evolver provides guidance to the Observer, any task seen by the Observer during training must also be excluded from Evolver evaluation. This guarantees that all 37 Evolver test cases fall within the Observer's 63 held-out tasks.

\begin{figure}[htbp]
  \centering
  \includegraphics[width=0.5\columnwidth]{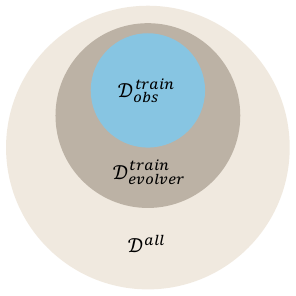}
  \caption{\textbf{Nested data partition structure.} $\mathcal{D}^{all}$ (86 tasks) contains $\mathcal{D}^{train}_{evolver}$ (49 successful trajectories from Claude Sonnet 4.5), which in turn contains $\mathcal{D}^{train}_{obs}$ (23 tasks covering 11 fault types). The strict subset relationship ensures that no test data appears in any training set. \emph{Visual interpretation:} The outer circle represents all benchmark tasks; the middle ring shows Evolver training data (all successful Sonnet trajectories); the innermost core shows Observer training data (carefully selected to cover diverse fault types while maintaining strict hold-out). This design enables fair evaluation of the combined Observer+Evolver system.}
  \label{fig:data-split}
\end{figure}

Figure~\ref{fig:data-split} and Tables~\ref{tab:train-fault-types}--\ref{tab:test-fault-types} detail the fault-type-based partition used for Observer GRPO training.

\begin{table}[htbp]
\centering
\caption{Training fault types (11 types $\to$ 23 successful trajectories). These fault types were selected to provide diverse coverage of common failure modes while ensuring sufficient test data remains for evaluation.}
\label{tab:train-fault-types}
\small
\begin{tabular}{lcc}
\toprule
\textbf{Fault Type} & \textbf{GT Tasks} & \textbf{Total Tasks} \\
\midrule
k8s\_target\_port-misconfig & 9 & 12 \\
scale\_pod\_zero\_social\_net & 3 & 4 \\
misconfig\_app\_hotel\_res & 3 & 4 \\
astronomy\_shop\_cart\_service\_failure & 1 & 2 \\
astronomy\_shop\_payment\_service\_failure & 1 & 2 \\
astronomy\_shop\_product\_catalog\_failure & 1 & 2 \\
astronomy\_shop\_recommend\_cache\_failure & 1 & 2 \\
auth\_miss\_mongodb & 1 & 4 \\
network\_loss\_hotel\_res & 1 & 2 \\
noop\_detection\_hotel\_reservation & 1 & 1 \\
redeploy\_without\_PV & 1 & 3 \\
\midrule
\textbf{Total} & \textbf{23} & \textbf{38} \\
\bottomrule
\end{tabular}
\end{table}

\begin{table}[htbp]
\centering
\caption{Test fault types (15 types $\to$ 48 tasks) plus 15 tasks from training fault types where Sonnet failed, totaling 63 held-out tasks. \emph{Critical:} All test fault types are completely unseen during Observer training, ensuring we evaluate true generalization capability.}
\label{tab:test-fault-types}
\small
\begin{tabular}{lc}
\toprule
\textbf{Fault Type} & \textbf{Tasks} \\
\midrule
revoke\_auth\_mongodb & 8 \\
user\_unregistered\_mongodb & 8 \\
assign\_to\_non\_existent\_node\_social\_net & 4 \\
wrong\_bin\_usage & 4 \\
astronomy\_shop\_ad\_service\_high\_cpu & 2 \\
astronomy\_shop\_ad\_service\_manual\_gc & 2 \\
astronomy\_shop\_ad\_service\_failure & 2 \\
network\_delay\_hotel\_res & 2 \\
pod\_kill\_hotel\_res & 2 \\
pod\_failure\_hotel\_res & 2 \\
astronomy\_shop\_image\_slow\_load & 2 \\
astronomy\_shop\_kafka\_queue\_problems & 2 \\
astronomy\_shop\_loadgen\_flood\_homepage & 2 \\
astronomy\_shop\_payment\_unreachable & 2 \\
noop\_detection\_social\_network & 1 \\
noop\_detection\_astronomy\_shop & 1 \\
container\_kill & 2 \\
\midrule
\emph{Test-only fault types subtotal} & \emph{48} \\
\emph{+ Training fault types (Sonnet failed)} & \emph{15} \\
\midrule
\textbf{Total held-out} & \textbf{63} \\
\bottomrule
\end{tabular}
\end{table}

\section{Multi-Round Sampling Analysis}
\label{app:rounds}

\noindent\textit{Organization note:} This section analyzes the stochastic behavior of the AOI system across multiple independent runs. We separate this from the experimental setup (Section B) because it focuses on \emph{empirical characterization} of the system's output distribution rather than configuration details. Understanding variance across runs is critical for deployment planning in production environments.

\subsection{Diminishing Returns in Sampling}

\begin{figure}[htbp]
  \centering
  \includegraphics[width=0.85\columnwidth]{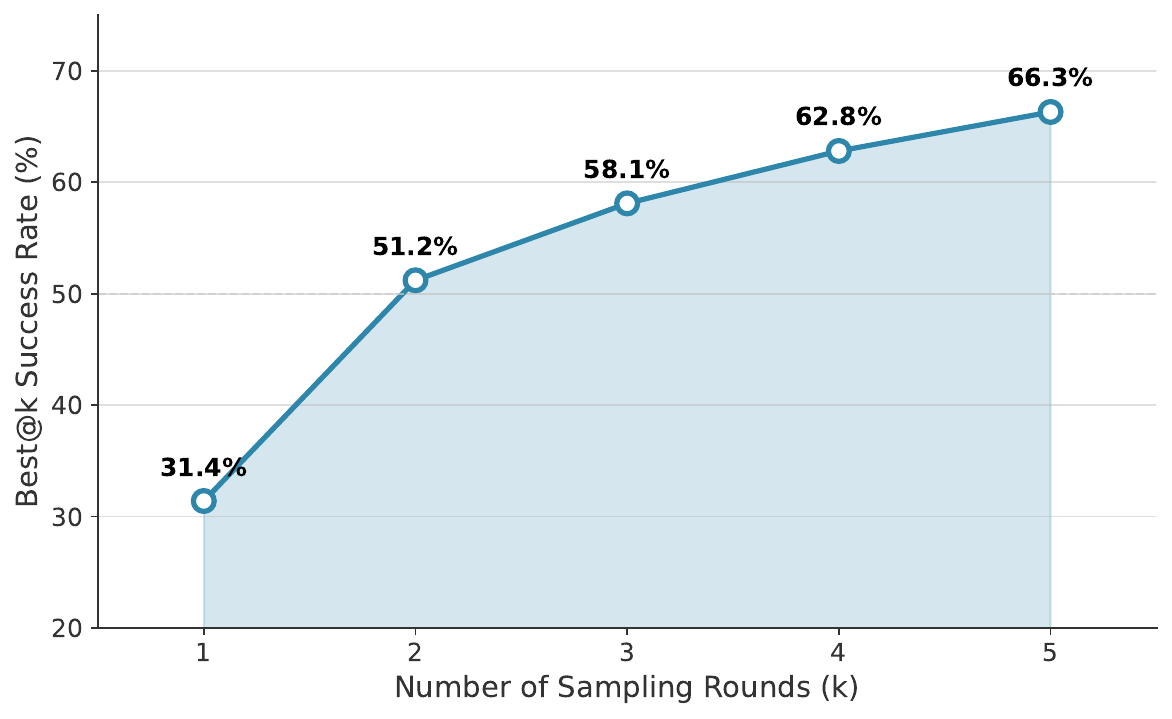}
  \caption{\textbf{best@$k$ success rate vs number of runs.} The curve shows clear diminishing returns: the first retry ($k$=1$\to$2) provides +19.8pp improvement, while subsequent retries ($k$=2$\to$5) each provide only +2-3pp. \emph{Interpretation:} The rapid initial improvement suggests many tasks have ``near-miss'' failure modes that succeed on retry, while the flattening curve indicates some tasks are fundamentally beyond current capability. The inflection point at $k$=2-3 provides clear guidance for resource-constrained deployments.}
  \label{fig:bestk}
\end{figure}

Figure~\ref{fig:bestk} shows success rate improvement across multiple sampling rounds.

\textbf{Practical deployment guidance.} The steep improvement from best@1 (31.4\%) to best@2 (51.2\%) suggests that even a single retry captures most of the ``easy wins'' from stochastic variation. Beyond $k$=3, gains become marginal (best@3=58.1\% $\to$ best@5=66.3\%). For cost-sensitive deployments, we recommend $k$=2 as the optimal trade-off; for high-stakes scenarios where maximizing success rate justifies additional compute, $k$=3 captures 88\% of the maximum achievable improvement.

\subsection{Per-Round Success Rates}

Table~\ref{tab:perround} shows success rates for each of the 5 independent runs, broken down by task type.

\begin{table}[htbp]
\centering
\caption{Per-round success rates (\%) by task type.}
\label{tab:perround}
\small
\begin{tabular}{lccccc}
\toprule
\textbf{Round} & \textbf{Det.} & \textbf{Loc.} & \textbf{RCA} & \textbf{Mit.} & \textbf{Overall} \\
\midrule
R1 & 53.1 & 21.4 & 7.7 & 23.1 & 31.4 \\
R2 & 75.0 & 32.1 & 0.0 & 23.1 & 41.9 \\
R3 & 59.4 & 32.1 & 15.4 & 23.1 & 38.4 \\
R4 & 75.0 & 21.4 & 15.4 & 23.1 & 40.7 \\
R5 & 71.9 & 32.1 & 0.0 & 23.1 & 40.7 \\
\midrule
Avg & 66.9 & 27.9 & 7.7 & 23.1 & 38.6 \\
\bottomrule
\end{tabular}
\end{table}

\textbf{Observations:} Detection shows high variance (53--75\%), indicating significant benefit from retries. Mitigation is perfectly stable (23.1\% every round), suggesting that failures are capability-limited rather than stochastic. RCA is highly unstable (0--15.4\%) due to small sample size (13 tasks) and complex multi-step reasoning requirements.

\begin{figure}[htbp]
  \centering
  \includegraphics[width=0.9\columnwidth]{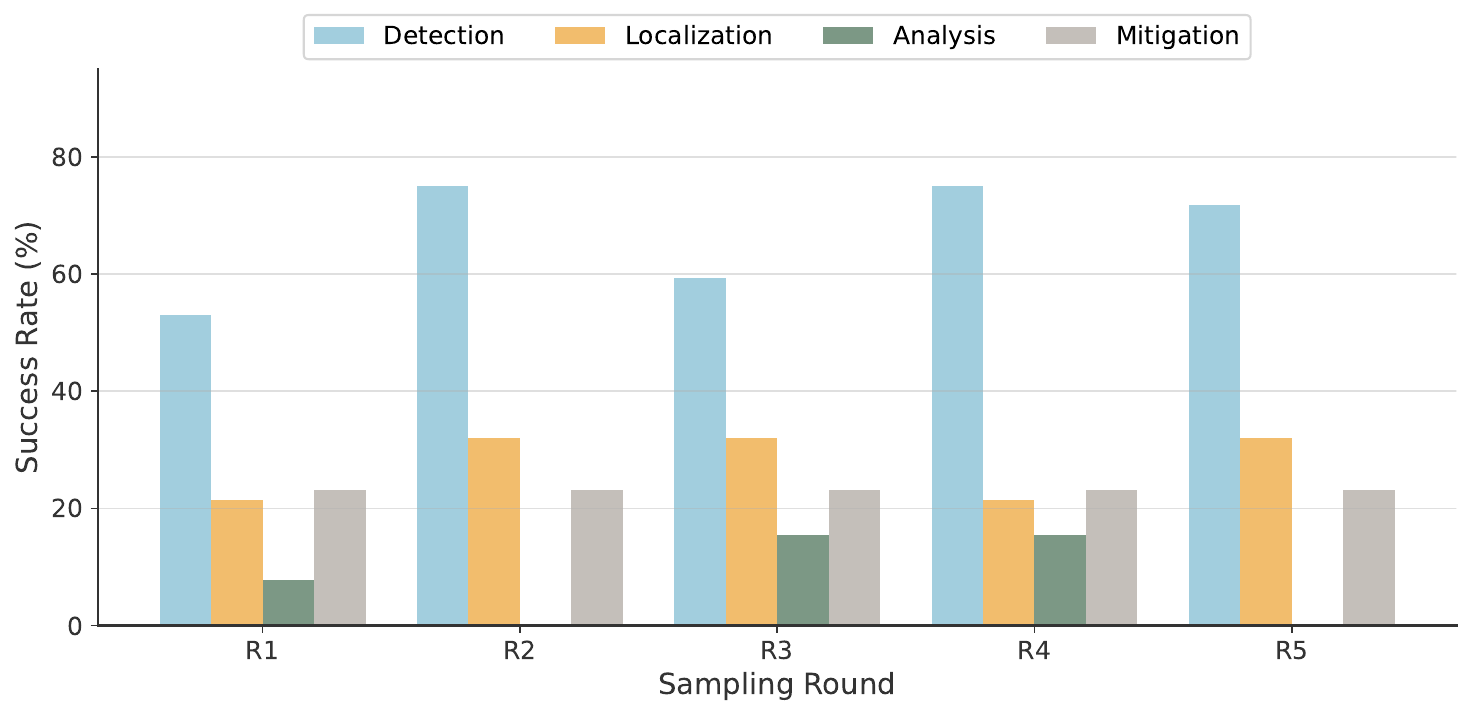}
  \caption{\textbf{Per-round success rate variance by task type.} Detection exhibits the widest range (22pp), indicating high stochastic sensitivity---tasks that fail in one run often succeed in another due to different exploration paths. Mitigation's flat line confirms that current failures are deterministic capability gaps rather than sampling artifacts. \emph{Actionable insight:} Detection tasks benefit most from multi-round sampling, while Mitigation requires architectural improvements rather than more attempts.}
  \label{fig:perround}
\end{figure}

Figure~\ref{fig:perround} visualizes this variance.

\subsection{Task Stability Distribution}

We categorize tasks by their 5-round success count to understand task difficulty distribution (Table~\ref{tab:stability}).

\begin{table}[htbp]
\centering
\caption{Task stability distribution (86 tasks total).}
\label{tab:stability}
\small
\begin{tabular}{llrr}
\toprule
\textbf{Success} & \textbf{Category} & \textbf{Count} & \textbf{\%} \\
\midrule
5/5 & Consistently solved & 14 & 16.3 \\
3--4/5 & Mostly solved & 16 & 18.6 \\
1--2/5 & Occasionally solved & 27 & 31.4 \\
0/5 & Never solved & 29 & 33.7 \\
\bottomrule
\end{tabular}
\end{table}

\textbf{Key insight:} 50\% of tasks (1--4/5) exhibit stochastic outcomes, strongly validating multi-round sampling. The 29 never-solved tasks represent hard capability boundaries requiring architectural improvements rather than more sampling. The 14 always-solved tasks demonstrate the system's reliable core competencies.

\begin{figure}[htbp]
  \centering
  \includegraphics[width=0.7\columnwidth]{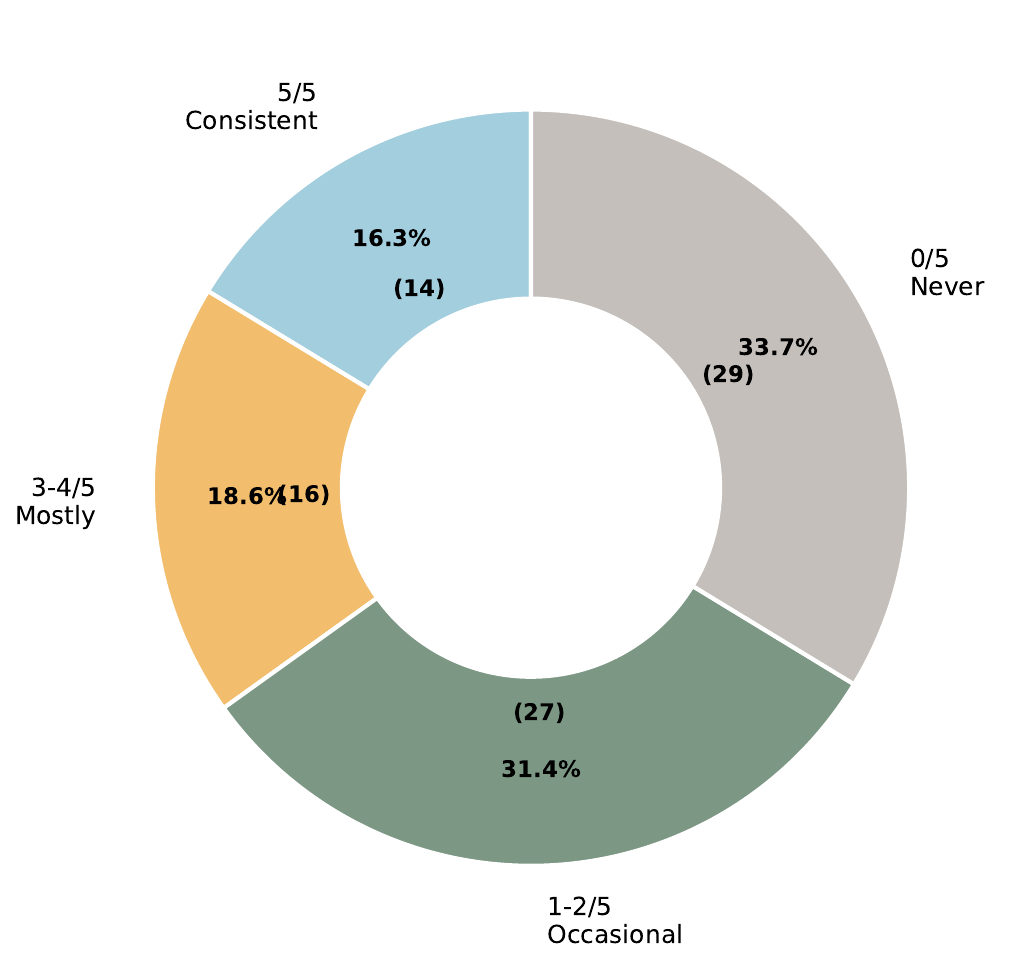}
  \caption{\textbf{Task stability distribution across 86 benchmark tasks.} The large middle segment (50\% of tasks with 1-4/5 success) validates our multi-round sampling strategy---these tasks are neither trivially easy nor impossibly hard, but rather exhibit outcome variance across runs. \emph{Design implication:} Future work should focus on converting the 0/5 category (capability gaps) into the 1-2/5 category (stochastic successes), which can then be reliably solved via multi-round sampling.}
  \label{fig:stability}
\end{figure}

Figure~\ref{fig:stability} visualizes this distribution.

\subsection{Never-Solved Tasks (0/5)}

The 29 consistently failed tasks cluster by type:
\begin{itemize}[leftmargin=*,topsep=2pt,itemsep=1pt]
\item \textbf{Localization (13):} Primarily astronomy\_shop microservice dependencies requiring deep trace analysis beyond current model capabilities
\item \textbf{RCA (9):} Causal reasoning tasks requiring understanding of temporal fault propagation
\item \textbf{Mitigation (7):} Domain-specific remediation commands (e.g., MongoDB auth recovery, Helm chart upgrades) not covered in training data
\end{itemize}

\section{Observer GRPO Analysis}
\label{app:grpo}

\noindent\textit{Organization note:} This section shifts from characterizing system behavior (Section C) to analyzing the \emph{effect of GRPO training}. We group all GRPO-related analysis together (Sections D and E) to provide a cohesive view of how training improves (and sometimes degrades) performance across different task types.

\subsection{Overall Training Effect}

\begin{figure}[htbp]
  \centering
  \includegraphics[width=0.9\columnwidth]{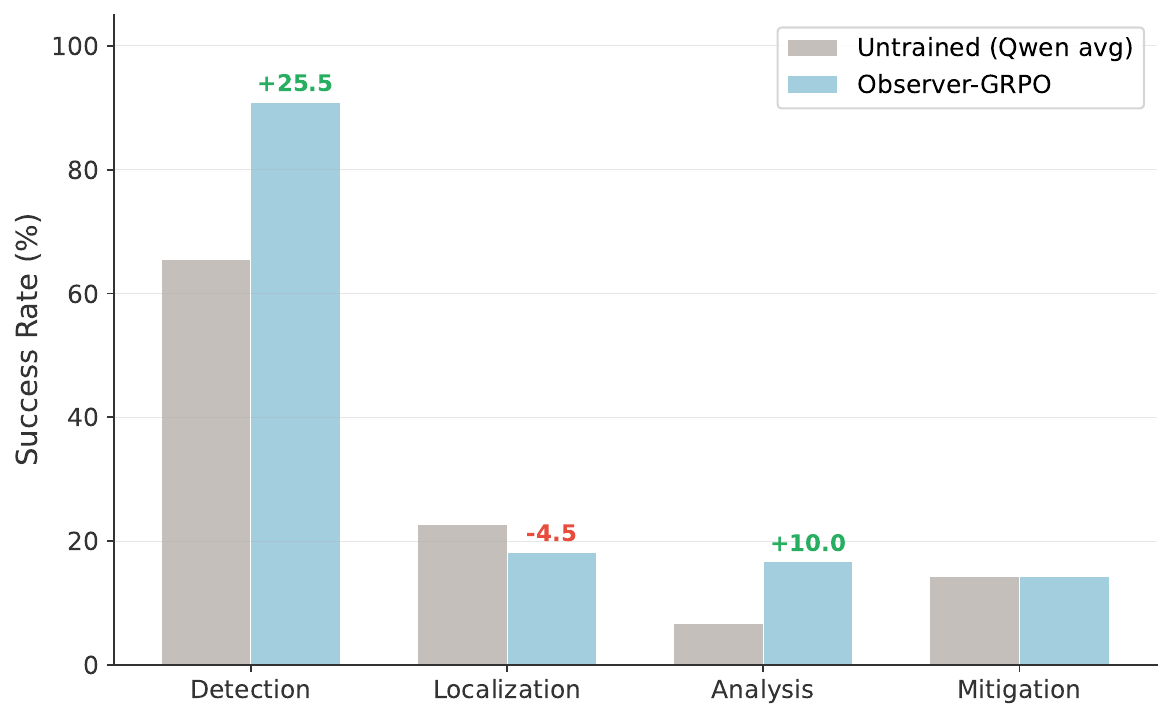}
  \caption{\textbf{Observer GRPO training effect by task type (best@5).} Detection improves dramatically (+25.5pp) as the Observer learns systematic anomaly-detection patterns. Localization \emph{decreases} ($-$4.5pp) due to over-exploration causing multi-anomaly confusion (detailed in Section~\ref{app:loc-analysis}). RCA improves (+8.4pp) from better evidence-gathering strategies. \emph{Critical insight:} GRPO training is not uniformly beneficial---it helps exploration-heavy tasks but can harm precision-critical tasks where thoroughness conflicts with efficiency.}
  \label{fig:grpo}
\end{figure}

Figure~\ref{fig:grpo} shows the aggregate impact of Observer GRPO training across task types.

\subsection{Task-Level Changes}

\begin{figure}[htbp]
  \centering
  \includegraphics[width=0.7\columnwidth]{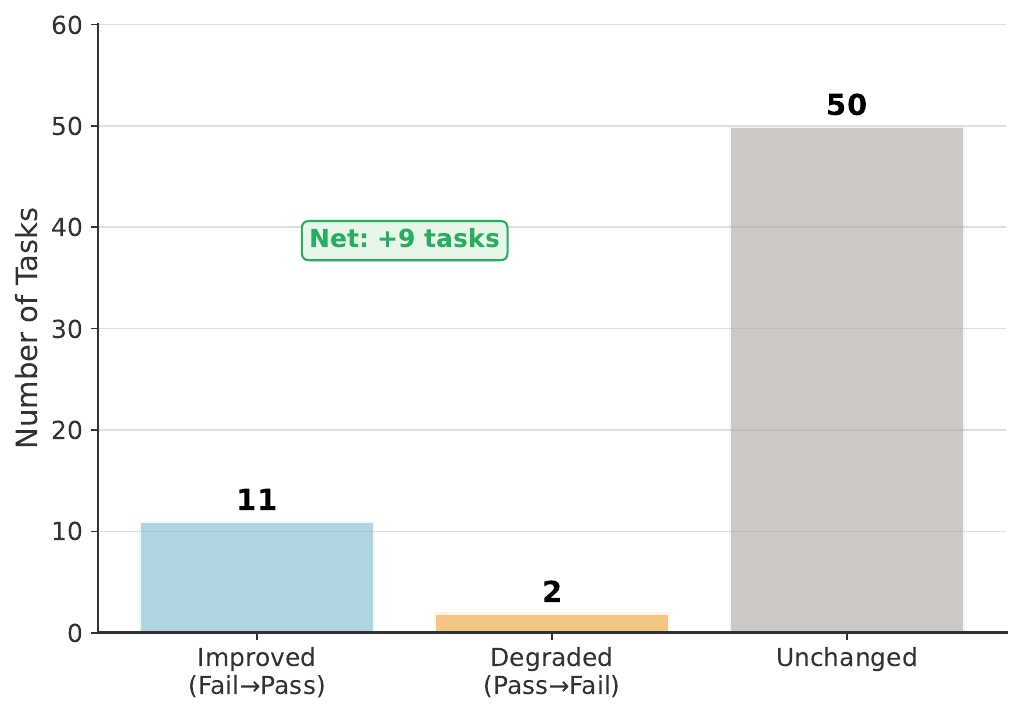}
  \caption{\textbf{Task-level changes after GRPO training.} Net +9 tasks improved (11 gains, 2 losses). Both degraded tasks are Localization, confirming that GRPO's deep-exploration strategy harms precision-critical fault localization. \emph{Interpretation:} The asymmetric gains (strong Detection improvements, Localization degradations) suggest task-type-aware training as future work.}
  \label{fig:grpo_changes}
\end{figure}

Figure~\ref{fig:grpo_changes} shows how individual tasks changed after GRPO training.

\subsection{Improved Tasks (11)}

Tasks that changed from mostly-failing (Qwen $<$50\%) to successful after GRPO:

\begin{itemize}[leftmargin=*,topsep=2pt,itemsep=1pt]
\item \textbf{Detection (7):} payment\_service\_unreachable, ad\_service\_failure, wrong\_bin\_usage, kafka\_queue\_problems, redeploy\_without\_PV, cart\_service\_failure, network\_delay\_hotel\_res
\item \textbf{Localization (2):} container\_kill, network\_delay\_hotel\_res
\item \textbf{RCA (2):} k8s\_target\_port-misconfig-analysis-2, revoke\_auth\_mongodb-analysis-2
\end{itemize}

\subsection{Degraded Tasks (2)}

Both degraded tasks are Localization:

\begin{itemize}[leftmargin=*,topsep=2pt,itemsep=1pt]
\item pod\_failure\_hotel\_res-localization-1 (4/5 $\to$ 0/5)
\item product\_catalog\_service\_failure-localization-1 (4/5 $\to$ 0/5)
\end{itemize}

\textbf{Root cause analysis.} Both degraded tasks require identifying a specific faulty component among multiple candidates with similar symptoms. Before GRPO, the base model would conservatively check only the most obvious candidate; after GRPO, the Observer learns to explore extensively---a strategy that succeeds for Detection (where finding \emph{any} issue suffices) but fails for Localization (where the \emph{exact} root cause must be identified). This reveals a fundamental tension: GRPO's reward signal optimizes for task completion, which can conflict with the thoroughness required for precise component identification.

\section{Observer-GRPO vs Base: Comprehensive Task-Type Analysis}
\label{app:rl-analysis}

This section provides detailed analysis of how GRPO training affects performance across different task types, revealing a critical trade-off between exploration depth and task requirements.

\subsection{Training Set Distribution}

The Observer GRPO training uses 23 successful trajectories with the following task-type distribution:

\begin{table}[htbp]
\centering
\caption{Training set composition by task type. The heavy Detection bias (43.5\%) explains why Detection benefits most from GRPO.}
\label{tab:train-dist}
\small
\begin{tabular}{lcc}
\toprule
\textbf{Task Type} & \textbf{Training Tasks} & \textbf{Test Tasks} \\
\midrule
Detection & 10 (43.5\%) & 22 \\
Localization & 6 (26.1\%) & 22 \\
Analysis (RCA) & 1 (4.3\%) & 12 \\
Mitigation & 6 (26.1\%) & 7 \\
\midrule
\textbf{Total} & \textbf{23} & \textbf{63} \\
\bottomrule
\end{tabular}
\end{table}

\textbf{Critical observation:} Analysis tasks are severely underrepresented in training (only 1 task, 4.3\%), yet Observer-GRPO shows large improvement on Analysis (+16.7pp). This suggests the learned exploration patterns transfer well despite limited direct supervision.

\subsection{Performance Comparison Overview}

\begin{figure}[htbp]
  \centering
  \includegraphics[width=0.85\columnwidth]{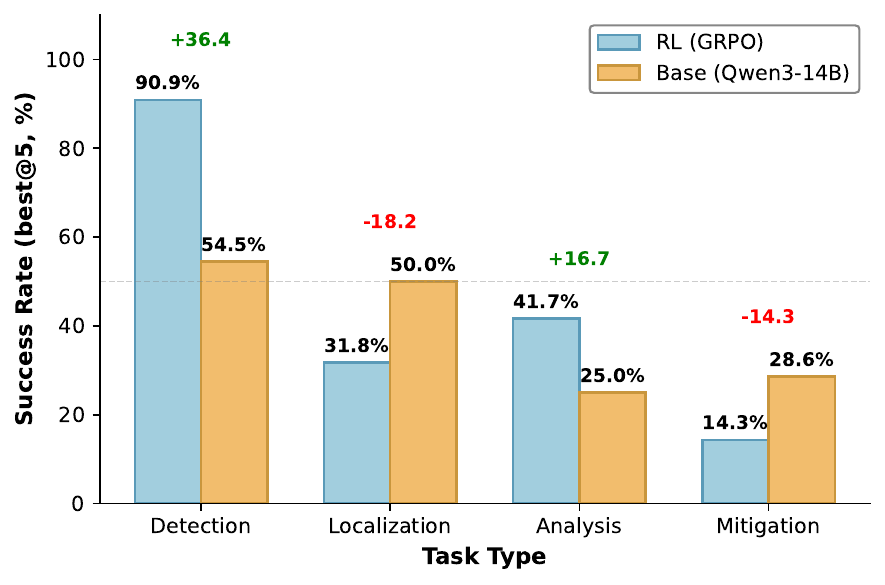}
  \caption{\textbf{Observer-GRPO vs Base performance by task type (best@5).} Detection improves by +36.4pp (from 59.1\% to 95.5\%), while Localization \emph{degrades} by $-$18.2pp (50.0\% to 31.8\%). \emph{Key takeaway:} GRPO benefits evidence-gathering tasks (Detection, Analysis) but harms precision tasks (Localization) where over-exploration introduces confusion. This heterogeneity motivates task-type-aware exploration policies.}
  \label{fig:rl-vs-base}
\end{figure}

Figure~\ref{fig:rl-vs-base} visualizes the performance gap between Observer-GRPO and Base models.

\subsection{Exploration Depth Analysis}

Table~\ref{tab:steps} compares the average number of exploration steps between Observer-GRPO and Base models.

\begin{table}[htbp]
\centering
\caption{Average exploration steps by task type (lower = more efficient).}
\label{tab:steps}
\small
\begin{tabular}{lccc}
\toprule
\textbf{Task Type} & \textbf{Obs-GRPO} & \textbf{Base} & \textbf{Diff.} \\
\midrule
Detection & 11.0 & 1.9 & +9.1 \\
Localization & 11.2 & 1.8 & +9.4 \\
Analysis & 9.6 & 1.1 & +8.5 \\
Mitigation & 12.0 & 2.8 & +9.2 \\
\midrule
\textbf{Average} & \textbf{10.9} & \textbf{1.9} & \textbf{+9.0} \\
\bottomrule
\end{tabular}
\end{table}

Observer-GRPO consistently uses $\sim$9 more steps than Base across all task types. This ``deep exploration'' strategy has task-type-dependent effects:

\begin{itemize}[leftmargin=*,topsep=2pt,itemsep=1pt]
\item \textbf{Detection}: \emph{Beneficial}---deeper search finds anomalies Base misses
\item \textbf{Localization}: \emph{Harmful}---over-exploration discovers multiple anomalies, causing incorrect root cause selection
\item \textbf{Analysis}: \emph{Beneficial}---deeper exploration gathers sufficient evidence for correct fault classification
\item \textbf{Mitigation}: Neutral effect---both struggle with execution-heavy tasks
\end{itemize}

\begin{figure}[htbp]
  \centering
  \includegraphics[width=0.8\columnwidth]{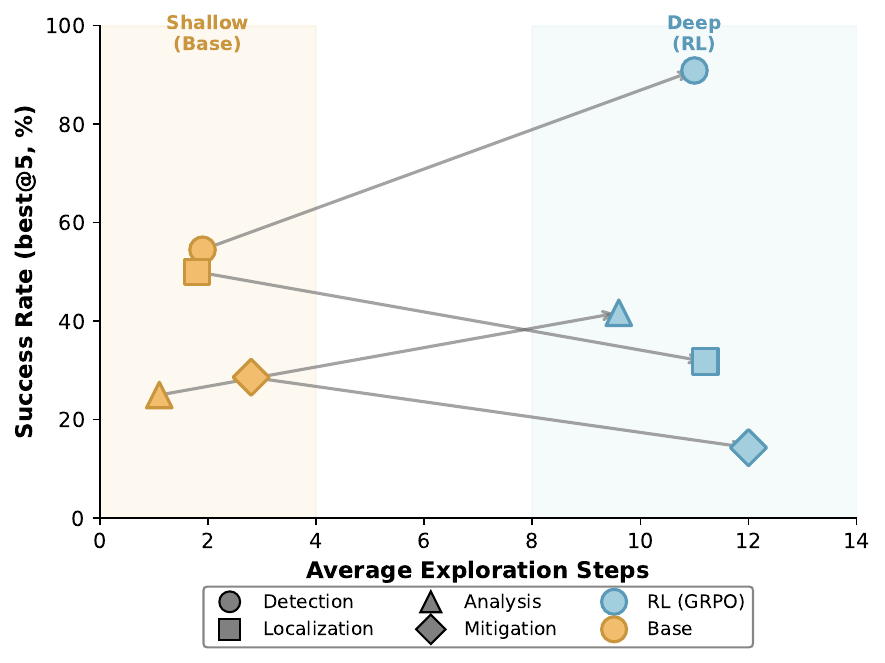}
  \caption{\textbf{Exploration depth vs success rate trade-off.} Each point represents a task type; arrows show the transition from Base to Observer-GRPO. GRPO uniformly increases exploration depth (arrows point right), but success direction varies: upward for Detection/Analysis (more exploration helps), downward for Localization (more exploration harms). \emph{Critical implication:} A single exploration strategy cannot optimize all task types simultaneously---future work should adapt exploration depth based on detected task category.}
  \label{fig:exploration-tradeoff}
\end{figure}

Figure~\ref{fig:exploration-tradeoff} visualizes the relationship between exploration depth and success rate.

\subsection{Localization: Why Observer-GRPO Underperforms}
\label{app:loc-analysis}

Table~\ref{tab:loc-detail} shows the detailed comparison for Localization tasks.

\begin{table}[htbp]
\centering
\caption{Localization task outcomes (22 tasks total).}
\label{tab:loc-detail}
\small
\begin{tabular}{lcc}
\toprule
\textbf{Outcome} & \textbf{Count} & \textbf{\%} \\
\midrule
Obs-GRPO wins (Base fails) & 2 & 9.1\% \\
Base wins (Obs-GRPO fails) & 6 & 27.3\% \\
Both succeed & 5 & 22.7\% \\
Both fail & 9 & 40.9\% \\
\midrule
\textbf{best@5} & \multicolumn{2}{c}{Obs-GRPO: 31.8\% vs Base: 50.0\%} \\
\bottomrule
\end{tabular}
\end{table}

\noindent{\small\textit{Obs-GRPO wins:} container\_kill, network\_loss\_hotel\_res.}\\
{\small\textit{Base wins:} ad\_service\_failure, ad\_service\_high\_cpu, payment\_service\_failure, k8s\_misconfig-loc-2, network\_delay, user\_unregistered-loc-2.}

\textbf{Root cause of Observer-GRPO degradation:} In the 6 tasks where Base wins:
\begin{enumerate}[leftmargin=*,topsep=2pt,itemsep=1pt]
\item Observer-GRPO explores extensively ($\sim$10 steps), discovering multiple components with anomalies
\item When multiple anomalies exist, Observer-GRPO often submits the \emph{symptom} rather than the \emph{root cause}
\item Base's shallow exploration (1--2 steps) paradoxically helps by limiting the search to the most obvious fault
\end{enumerate}

\textbf{Case study: astronomy\_shop\_ad\_service\_failure-localization-1}
\begin{itemize}[leftmargin=*,topsep=2pt,itemsep=1pt]
\item \textbf{Ground truth}: The \texttt{ad} service is the root cause
\item \textbf{Base} (2 steps): Quickly identifies \texttt{ad} service errors $\to$ submits \texttt{["ad"]} $\to$ \textcolor{green!50!black}{\textbf{Correct}}
\item \textbf{Obs-GRPO} (10 steps): Finds \texttt{ad} errors, explores dependencies, finds \texttt{product-catalog} affected $\to$ submits \texttt{["product-catalog"]} $\to$ \textcolor{red}{\textbf{Incorrect}} (downstream symptom, not root cause)
\end{itemize}

\subsection{Analysis: Why Observer-GRPO Excels}
\label{app:rca-analysis}

Table~\ref{tab:rca-detail} shows the detailed comparison for Analysis (RCA) tasks.

\begin{table}[htbp]
\centering
\caption{Analysis task outcomes (12 tasks total).}
\label{tab:rca-detail}
\small
\begin{tabular}{lcc}
\toprule
\textbf{Outcome} & \textbf{Count} & \textbf{\%} \\
\midrule
Obs-GRPO wins (Base fails) & 2 & 16.7\% \\
Base wins (Obs-GRPO fails) & 0 & 0.0\% \\
Both succeed & 3 & 25.0\% \\
Both fail & 7 & 58.3\% \\
\midrule
\textbf{best@5} & \multicolumn{2}{c}{Obs-GRPO: 41.7\% vs Base: 25.0\%} \\
\bottomrule
\end{tabular}
\end{table}

\noindent{\small\textit{Obs-GRPO wins:} k8s\_misconfig-analysis-1, revoke\_auth-analysis-1.}\\
{\small\textit{Both succeed:} k8s\_misconfig-analysis-2, misconfig\_app, revoke\_auth-analysis-2.}

\textbf{Why deeper exploration helps Analysis:}
Analysis tasks require determining both \texttt{system\_level} (e.g., Application, Virtualization, Network) and \texttt{fault\_type} (e.g., Misconfiguration, Resource Exhaustion). This classification requires:
\begin{enumerate}[leftmargin=*,topsep=2pt,itemsep=1pt]
\item Understanding which layer the fault originates from
\item Gathering evidence about the nature of the fault
\item Eliminating alternative hypotheses
\end{enumerate}

Base's shallow exploration (avg 1.1 steps) provides insufficient evidence for reliable classification. Observer-GRPO's deeper exploration (avg 9.6 steps) systematically gathers the necessary information.

\textbf{Case study: k8s\_target\_port-misconfig-analysis-1}
\begin{itemize}[leftmargin=*,topsep=2pt,itemsep=1pt]
\item \textbf{Ground truth}: \texttt{system\_level: Virtualization, fault\_type: Misconfiguration}
\item \textbf{Base} (1.6 steps avg): Insufficient exploration $\to$ 0/5 rounds correct
\item \textbf{Obs-GRPO} (10 steps): Examines service definitions, finds \texttt{targetPort} mismatch $\to$ identifies Virtualization Misconfiguration $\to$ 4/5 correct
\end{itemize}

\subsection{Task Outcome Distribution}

\begin{figure}[htbp]
  \centering
  \includegraphics[width=0.95\columnwidth]{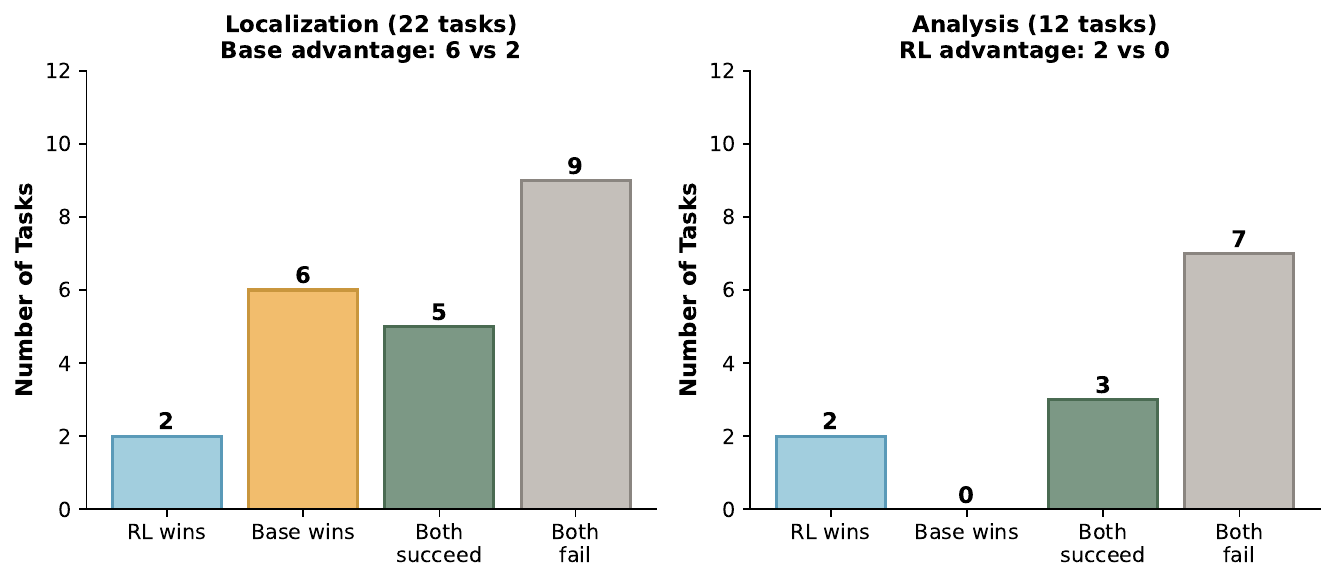}
  \caption{\textbf{Task outcome distribution for Localization vs Analysis.} Localization shows Base advantage (6 exclusive wins vs 2), while Analysis shows Obs-GRPO advantage (2 vs 0). \emph{Interpretation:} Confirms that deep exploration helps Analysis (which benefits from comprehensive evidence) but harms Localization (where first instinct is often correct). The asymmetry validates task-type-specific exploration strategies.}
  \label{fig:task-outcomes}
\end{figure}

Figure~\ref{fig:task-outcomes} compares task-level outcomes between Observer-GRPO and Base for Localization and Analysis.

\subsection{The Exploration-Precision Trade-off}

Our analysis reveals a fundamental trade-off in diagnostic agent design:

\begin{table}[htbp]
\centering
\caption{Exploration depth trade-off by task type.}
\label{tab:tradeoff}
\small
\begin{tabular}{lcc}
\toprule
\textbf{Task Type} & \textbf{Optimal Strategy} & \textbf{Why} \\
\midrule
Detection & Shallow OK & Any anomaly suffices \\
Localization & Shallow better & Avoid multi-anomaly confusion \\
Analysis & Deep better & Need comprehensive evidence \\
Mitigation & Deep required & Must understand before fixing \\
\bottomrule
\end{tabular}
\end{table}

\textbf{Implications for future work:}
\begin{enumerate}[leftmargin=*,topsep=2pt,itemsep=1pt]
\item \textbf{Task-aware exploration}: Dynamically adjust exploration depth based on detected task type
\item \textbf{Confidence-based early stopping}: For Localization, stop when a high-confidence root cause is found
\item \textbf{Multi-stage training}: Train separate policies for different task types, or use task-type-conditioned rewards
\end{enumerate}

\begin{figure}[htbp]
  \centering
  \includegraphics[width=0.6\columnwidth]{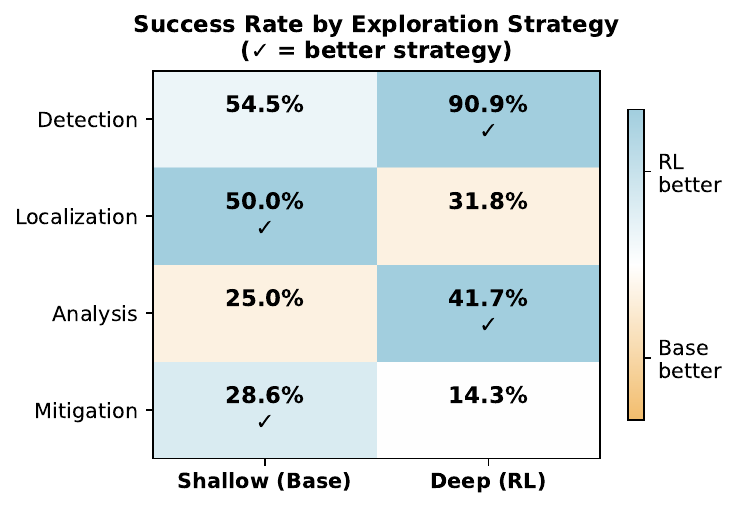}
  \caption{\textbf{Exploration strategy effectiveness heatmap.} Green = strategy helps; Red = strategy harms. Localization uniquely benefits from shallow exploration, while Analysis requires deep exploration. This heterogeneity strongly motivates task-type-aware policies as future work.}
  \label{fig:strategy-summary}
\end{figure}

Figure~\ref{fig:strategy-summary} provides a visual summary.

\section{Case Studies: Detailed Trajectory Examples}
\label{app:trajectory}

\noindent\textit{Organization note:} After presenting quantitative results (Sections C-E), this section provides \emph{qualitative illustration} through detailed execution traces. We include multiple case studies covering different task types and failure modes to demonstrate how the multi-agent architecture operates in practice. Each case study highlights specific design decisions (read-write separation, dual-timescale memory, Evolver guidance) in action.

\subsection{Case Study: Mitigation with Storage Class Failure}
\label{app:case1}

\textbf{Why this case:} This example illustrates the complete \method pipeline---Observer reasoning, Probe exploration, Executor remediation, and verification. It demonstrates: (1) how the Observer identifies root causes through systematic exploration; (2) how read-write separation prevents premature state mutations; (3) how the Executor applies fixes only after sufficient evidence accumulates.

\textbf{Scenario:} \texttt{redeploy\_without\_PV-mitigation-1}. The StorageClass for persistent volumes was removed during redeployment, causing cascading PVC binding failures and pod crashes.

\begin{tcolorbox}[colback=colorD!20,colframe=colorD!80!black,fontupper=\ttfamily\scriptsize,title={\small Phase: Initialization \hfill Run 1}]
{[}04:00:28{]} Fault injected: StorageClass `local-storage' deleted\\
{[}04:00:29{]} Namespace `test-hotel-reservation' redeployed without PV\\
{[}04:00:35{]} Multiple pods entering CrashLoopBackOff state\\
Task Started
\end{tcolorbox}

\begin{tcolorbox}[colback=colorA!20,colframe=colorA!80!black,title={\small Thought \& Action \hfill Observer Agent}]
\small
The task requires mitigating anomalies in test-hotel-reservation. I should first understand the current state of the system by listing all resources in the namespace.

 \hrule 

\textbf{Action:} Probe\\
\textbf{Subtask:} ``List all pods/services in namespace to identify failing components''
\end{tcolorbox}

\begin{tcolorbox}[colback=colorB!20,colframe=colorB!80!black,fontupper=\ttfamily\scriptsize,title={\small Tool Usage: kubectl \hfill Probe Agent}]
\$ kubectl get pods -n test-hotel-reservation\\
NAME \hspace{3.5cm} READY \hspace{0.3cm} STATUS \hspace{1.5cm} RESTARTS\\
consul-b84877fb4-v9sgv \hspace{0.6cm} 1/1 \hspace{0.5cm} Running \hspace{1.2cm} 0\\
geo-99d8c58c-5wrbn \hspace{1.1cm} 0/1 \hspace{0.5cm} \textcolor{red}{CrashLoopBackOff} \hspace{0.1cm} 3\\
mongodb-geo-5ff578bcb8 \hspace{0.55cm} 0/1 \hspace{0.5cm} \textcolor{red}{Pending} \hspace{1.3cm} 0\\
mongodb-profile-758cb \hspace{0.7cm} 0/1 \hspace{0.5cm} \textcolor{red}{Pending} \hspace{1.3cm} 0\\
reservation-78c9f79 \hspace{1.05cm} 0/1 \hspace{0.5cm} \textcolor{red}{CrashLoopBackOff} \hspace{0.1cm} 5\\
... (8 more pods in error state)
\end{tcolorbox}

\begin{tcolorbox}[colback=colorA!20,colframe=colorA!80!black,title={\small Thought \& Action \hfill Observer Agent}]
\small
Multiple pods are failing. The mongodb pods are in Pending state while service pods are in CrashLoopBackOff. This suggests a volume mounting issue. I need to check the PVC status.

 \hrule 

\textbf{Action:} Probe\\
\textbf{Subtask:} ``Check PVC status and describe pending pods''
\end{tcolorbox}
\begin{tcolorbox}[colback=colorB!20,colframe=colorB!80!black,fontupper=\ttfamily\scriptsize,title={\small Tool Usage: kubectl \hfill Probe Agent}]
\$ kubectl get pvc -n test-hotel-reservation\\
NAME \hspace{2cm} STATUS \hspace{0.5cm} STORAGECLASS\\
mongodb-geo-pvc \hspace{0.5cm} \textcolor{red}{Pending} \hspace{0.3cm} local-storage\\
mongodb-profile-pvc \hspace{0.2cm} \textcolor{red}{Pending} \hspace{0.3cm} local-storage\\
mongodb-rate-pvc \hspace{0.45cm} \textcolor{red}{Pending} \hspace{0.3cm} local-storage\\
\\
\$ kubectl describe pvc mongodb-geo-pvc -n test-hotel-reservation\\
Events:\\
  Warning \hspace{0.3cm} ProvisioningFailed \hspace{0.3cm} \textcolor{red}{storageclass.storage.k8s.io}\\
  \hspace{1.5cm} \textcolor{red}{`local-storage' not found}
\end{tcolorbox}
\begin{tcolorbox}[colback=colorA!20,colframe=colorA!80!black,title={\small Thought \& Action \hfill Observer Agent}]
\small
Root cause identified: The StorageClass ``local-storage'' is missing, preventing PVCs from binding. The mongodb pods cannot start without their persistent volumes, and dependent services fail.
\textbf{Solution:} Recreate the StorageClass, then trigger PVC rebinding.
 \hrule 
\textbf{Action:} Execute\\
\textbf{Instruction:} ``Create StorageClass 'local-storage' with provisioner 'kubernetes.io/no-provisioner' and volumeBindingMode 'WaitForFirstConsumer'''
\end{tcolorbox}
\begin{tcolorbox}[colback=colorB!20,colframe=colorB!80!black,fontupper=\ttfamily\scriptsize,title={\small Tool Usage: kubectl apply \hfill Executor Agent}]
\$ kubectl apply -f storageclass.yaml\\
\\
apiVersion: storage.k8s.io/v1\\
kind: StorageClass\\
metadata: name: local-storage\\
provisioner: kubernetes.io/no-provisioner\\
\\
storageclass.storage.k8s.io/local-storage \textcolor{green!50!black}{created}
\end{tcolorbox}
\begin{tcolorbox}[colback=colorA!20,colframe=colorA!80!black,title={\small Thought \& Action \hfill Observer Agent}]
\small
StorageClass recreated. Now I need to delete the stuck PVCs and let them rebind, then restart the deployments to pick up the changes.
 \hrule 
\textbf{Action:} Execute\\
\textbf{Instruction:} ``Delete pending PVCs and restart all deployments''
\end{tcolorbox}
\begin{tcolorbox}[colback=colorB!20,colframe=colorB!80!black,fontupper=\ttfamily\scriptsize,title={\small Tool Usage: kubectl \hfill Executor Agent}]
\$ kubectl delete pvc --all -n test-hotel-reservation\\
persistentvolumeclaim `mongodb-geo-pvc' \textcolor{green!50!black}{deleted}\\
persistentvolumeclaim `mongodb-profile-pvc' \textcolor{green!50!black}{deleted}\\
persistentvolumeclaim `mongodb-rate-pvc' \textcolor{green!50!black}{deleted}\\
\\
\$ kubectl rollout restart deployment --all -n test-hotel-reservation\\
deployment.apps/consul \textcolor{green!50!black}{restarted}\\
deployment.apps/geo \textcolor{green!50!black}{restarted}\\
deployment.apps/mongodb-geo \textcolor{green!50!black}{restarted}\\
... (9 more deployments restarted)
\end{tcolorbox}
\begin{tcolorbox}[colback=colorA!20,colframe=colorA!80!black,title={\small Thought \& Action \hfill Observer Agent}]
\small
Mitigation commands executed. Let me verify that all pods are now running correctly before submitting.
 \hrule 
\textbf{Action:} Probe\\
\textbf{Subtask:} ``Verify all pods are Running''
\end{tcolorbox}
\begin{tcolorbox}[colback=colorB!20,colframe=colorB!80!black,fontupper=\ttfamily\scriptsize,title={\small Tool Usage: kubectl \hfill Probe Agent}]
\$ kubectl get pods -n test-hotel-reservation\\
NAME \hspace{3.5cm} READY \hspace{0.3cm} STATUS \hspace{0.5cm} RESTARTS\\
consul-b84877fb4-x2kpl \hspace{0.6cm} 1/1 \hspace{0.5cm} \textcolor{green!50!black}{Running} \hspace{0.3cm} 0\\
geo-99d8c58c-7mnqr \hspace{1.1cm} 1/1 \hspace{0.5cm} \textcolor{green!50!black}{Running} \hspace{0.3cm} 0\\
mongodb-geo-5ff578b-new \hspace{0.35cm} 1/1 \hspace{0.5cm} \textcolor{green!50!black}{Running} \hspace{0.3cm} 0\\
mongodb-profile-758cb-new \hspace{0.15cm} 1/1 \hspace{0.5cm} \textcolor{green!50!black}{Running} \hspace{0.3cm} 0\\
reservation-78c9f79-new \hspace{0.4cm} 1/1 \hspace{0.5cm} \textcolor{green!50!black}{Running} \hspace{0.3cm} 0\\
... (all 12 pods Running)
\end{tcolorbox}
\begin{tcolorbox}[colback=colorC!20,colframe=colorC!80!black,title={\small Task Finished \hfill Observer Agent}]
\small
\textbf{Diagnosis:} StorageClass ``local-storage'' was missing after redeployment, causing PVC binding failures and cascading pod failures.
\textbf{Mitigation:} (1) Recreated StorageClass with correct provisioner; (2) Deleted stuck PVCs to trigger rebinding; (3) Restarted all deployments to pick up new PVCs.
\textbf{Result:} All 12 pods now Running. Services operational.
\end{tcolorbox}
\begin{tcolorbox}[colback=colorD!20,colframe=colorD!80!black,fontupper=\ttfamily\scriptsize,title={\small Phase: Validation \hfill Run 1}]
{[}04:19:08{]} Validation started...\\
{[}04:19:10{]} Pod check: 12/12 Running \textcolor{green!50!black}{PASS}\\
{[}04:19:12{]} Service check: All endpoints healthy \textcolor{green!50!black}{PASS}\\
{[}04:19:15{]} \textcolor{green!50!black}{Task completed successfully}\\
Total time: 1139 seconds, 13 iterations
\end{tcolorbox}
   
\noindent\textbf{Key takeaways from this case:}
\begin{itemize}[leftmargin=*,topsep=2pt,itemsep=1pt]
\item \textbf{Systematic exploration:} Observer first gathers evidence (pod status, PVC status) before taking action
\item \textbf{Read-write separation:} All diagnostic commands execute through Probe; only after root cause is confirmed does Executor engage
\item \textbf{Verification loop:} After applying fixes, Observer verifies success before submitting
\item \textbf{Multi-step reasoning:} Recognizes that fixing StorageClass alone isn't enough---must also trigger PVC rebinding and deployment restart
\end{itemize}


\section{Detailed Per-Task Results}
\label{app:detailed}

\noindent\textit{Organization note:} This final section provides the complete empirical record of all 86 tasks across 5 runs. We place this at the end because it serves as a reference appendix rather than primary analysis. Researchers can use these tables to identify specific tasks for further investigation or to verify aggregate statistics reported in the main paper.

Tables~\ref{tab:detailed_det}--\ref{tab:detailed_mit} present the complete per-task results across 5 independent runs.
\textcolor{green!60!black}{\textbf{T}} indicates success, \textcolor{red!70!black}{\textbf{F}} indicates failure.

\colorlet{passcolor}{green!20}
\colorlet{failcolor}{red!20}
\newcommand{\pT}{\cellcolor{passcolor}\textbf{T}}
\newcommand{\pF}{\cellcolor{failcolor}\textbf{F}}

\begin{table}[H]
\centering
\caption{Detailed results: \textbf{Detection} tasks (32 total). High variance across runs indicates stochastic sensitivity to exploration paths.}
\label{tab:detailed_det}
\scriptsize
\setlength{\tabcolsep}{2pt}
\begin{tabular}{lcccccc}
\toprule
\textbf{Problem ID} & \textbf{R1} & \textbf{R2} & \textbf{R3} & \textbf{R4} & \textbf{R5} & \textbf{Best} \\
\midrule
assign\_to\_non\_existent\_node-det-1 & \pT & \pT & \pT & \pT & \pT & 5/5 \\
astronomy\_shop\_ad\_service\_failure-det-1 & \pF & \pT & \pF & \pT & \pF & 2/5 \\
astronomy\_shop\_ad\_service\_high\_cpu-det-1 & \pT & \pT & \pF & \pT & \pT & 4/5 \\
astronomy\_shop\_ad\_service\_manual\_gc-det-1 & \pF & \pT & \pT & \pT & \pF & 3/5 \\
astronomy\_shop\_cart\_service\_failure-det-1 & \pF & \pT & \pF & \pT & \pF & 2/5 \\
astronomy\_shop\_image\_slow\_load-det-1 & \pF & \pT & \pT & \pT & \pT & 4/5 \\
astronomy\_shop\_kafka\_queue\_problems-det-1 & \pF & \pT & \pT & \pF & \pF & 2/5 \\
astronomy\_shop\_loadgen\_flood-det-1 & \pT & \pT & \pT & \pT & \pT & 5/5 \\
astronomy\_shop\_payment\_failure-det-1 & \pF & \pT & \pT & \pT & \pF & 3/5 \\
astronomy\_shop\_payment\_unreachable-det-1 & \pF & \pT & \pF & \pT & \pF & 2/5 \\
astronomy\_shop\_product\_catalog-det-1 & \pT & \pT & \pT & \pT & \pT & 5/5 \\
astronomy\_shop\_recommend\_cache-det-1 & \pT & \pT & \pF & \pF & \pT & 3/5 \\
auth\_miss\_mongodb-det-1 & \pT & \pT & \pT & \pT & \pT & 5/5 \\
container\_kill-det & \pT & \pF & \pF & \pT & \pT & 3/5 \\
k8s\_target\_port-misconfig-det-1 & \pT & \pF & \pF & \pF & \pF & 1/5 \\
k8s\_target\_port-misconfig-det-2 & \pF & \pF & \pF & \pT & \pT & 2/5 \\
k8s\_target\_port-misconfig-det-3 & \pF & \pT & \pT & \pT & \pT & 4/5 \\
misconfig\_app\_hotel\_res-det-1 & \pT & \pT & \pT & \pT & \pT & 5/5 \\
network\_delay\_hotel\_res-det-1 & \pF & \pF & \pT & \pF & \pT & 2/5 \\
network\_loss\_hotel\_res-det-1 & \pF & \pF & \pF & \pF & \pT & 1/5 \\
noop\_detection\_astronomy\_shop-1 & \pT & \pF & \pF & \pF & \pT & 2/5 \\
noop\_detection\_hotel\_reservation-1 & \pT & \pT & \pT & \pF & \pT & 4/5 \\
noop\_detection\_social\_network-1 & \pF & \pT & \pT & \pT & \pT & 4/5 \\
pod\_failure\_hotel\_res-det-1 & \pT & \pT & \pT & \pT & \pT & 5/5 \\
pod\_kill\_hotel\_res-det-1 & \pF & \pF & \pF & \pF & \pT & 1/5 \\
redeploy\_without\_PV-det-1 & \pF & \pF & \pF & \pT & \pF & 1/5 \\
revoke\_auth\_mongodb-det-1 & \pT & \pT & \pT & \pT & \pT & 5/5 \\
revoke\_auth\_mongodb-det-2 & \pT & \pT & \pT & \pT & \pT & 5/5 \\
scale\_pod\_zero\_social\_net-det-1 & \pT & \pT & \pT & \pT & \pT & 5/5 \\
user\_unregistered\_mongodb-det-1 & \pT & \pT & \pT & \pT & \pT & 5/5 \\
user\_unregistered\_mongodb-det-2 & \pT & \pT & \pT & \pT & \pT & 5/5 \\
wrong\_bin\_usage-det-1 & \pF & \pT & \pF & \pT & \pF & 2/5 \\
\midrule
\textbf{Total} & 17/32 & 24/32 & 19/32 & 24/32 & 23/32 & -- \\
\bottomrule
\end{tabular}
\end{table}

  \begin{table}[H]
  \centering
  \caption{Detailed results: \textbf{Localization} tasks (28 total).}
  \label{tab:detailed_loc}
  \scriptsize
  \setlength{\tabcolsep}{2pt}
  \begin{tabular}{lcccccc}
  \toprule
  \textbf{Problem ID} & \textbf{R1} & \textbf{R2} & \textbf{R3} & \textbf{R4} & \textbf{R5} & \textbf{Best} \\
  \midrule
  assign\_to\_non\_existent\_node-loc-1 & \pT & \pT & \pT & \pT & \pT & 5/5 \\
  astronomy\_shop\_ad\_service\_failure-loc-1 & \pF & \pT & \pF & \pF & \pF & 1/5 \\
  astronomy\_shop\_ad\_service\_high\_cpu-loc-1 & \pF & \pT & \pF & \pF & \pT & 2/5 \\
  astronomy\_shop\_ad\_service\_manual\_gc-loc-1 & \pF & \pF & \pF & \pF & \pF & 0/5 \\
  astronomy\_shop\_cart\_service\_failure-loc-1 & \pF & \pF & \pF & \pF & \pF & 0/5 \\
  astronomy\_shop\_image\_slow\_load-loc-1 & \pF & \pF & \pF & \pF & \pF & 0/5 \\
  astronomy\_shop\_kafka\_queue\_problems-loc-1 & \pF & \pF & \pF & \pF & \pF & 0/5 \\
  astronomy\_shop\_loadgen\_flood-loc-1 & \pF & \pF & \pF & \pF & \pF & 0/5 \\
  astronomy\_shop\_payment\_failure-loc-1 & \pF & \pF & \pF & \pF & \pT & 1/5 \\
  astronomy\_shop\_payment\_unreachable-loc-1 & \pF & \pF & \pF & \pF & \pF & 0/5 \\
  astronomy\_shop\_product\_catalog-loc-1 & \pT & \pT & \pT & \pF & \pT & 4/5 \\
  astronomy\_shop\_recommend\_cache-loc-1 & \pF & \pF & \pF & \pF & \pF & 0/5 \\
  auth\_miss\_mongodb-loc-1 & \pF & \pF & \pT & \pT & \pT & 3/5 \\
  container\_kill-loc & \pF & \pF & \pF & \pF & \pF & 0/5 \\
  k8s\_target\_port-misconfig-loc-1 & \pT & \pF & \pT & \pT & \pF & 3/5 \\
  k8s\_target\_port-misconfig-loc-2 & \pF & \pT & \pF & \pF & \pT & 2/5 \\
  k8s\_target\_port-misconfig-loc-3 & \pT & \pT & \pT & \pF & \pF & 3/5 \\
  misconfig\_app\_hotel\_res-loc-1 & \pT & \pF & \pT & \pT & \pT & 4/5 \\
  network\_delay\_hotel\_res-loc-1 & \pF & \pT & \pF & \pF & \pF & 1/5 \\
  network\_loss\_hotel\_res-loc-1 & \pF & \pF & \pF & \pF & \pF & 0/5 \\
  pod\_failure\_hotel\_res-loc-1 & \pF & \pT & \pT & \pT & \pT & 4/5 \\
  pod\_kill\_hotel\_res-loc-1 & \pF & \pF & \pF & \pF & \pF & 0/5 \\
  revoke\_auth\_mongodb-loc-1 & \pT & \pF & \pF & \pF & \pF & 1/5 \\
  revoke\_auth\_mongodb-loc-2 & \pF & \pF & \pF & \pF & \pF & 0/5 \\
  scale\_pod\_zero\_social\_net-loc-1 & \pF & \pT & \pT & \pT & \pT & 4/5 \\
  user\_unregistered\_mongodb-loc-1 & \pF & \pF & \pF & \pF & \pF & 0/5 \\
  user\_unregistered\_mongodb-loc-2 & \pF & \pF & \pT & \pF & \pF & 1/5 \\
  wrong\_bin\_usage-loc-1 & \pF & \pF & \pF & \pF & \pF & 0/5 \\
  \midrule
  \textbf{Total} & 6/28 & 9/28 & 9/28 & 6/28 & 9/28 & -- \\
  \bottomrule
  \end{tabular}
  \end{table}

  \begin{table}[H]
  \centering
  \caption{Detailed results: \textbf{RCA} tasks (13 total).}
  \label{tab:detailed_rca}
  \scriptsize
  \setlength{\tabcolsep}{2pt}
  \begin{tabular}{lcccccc}
  \toprule
  \textbf{Problem ID} & \textbf{R1} & \textbf{R2} & \textbf{R3} & \textbf{R4} & \textbf{R5} & \textbf{Best} \\
  \midrule
  assign\_to\_non\_existent\_node-rca-1 & \pF & \pF & \pF & \pF & \pF & 0/5 \\
  auth\_miss\_mongodb-rca-1 & \pF & \pF & \pF & \pF & \pF & 0/5 \\
  k8s\_target\_port-misconfig-rca-1 & \pF & \pF & \pF & \pF & \pF & 0/5 \\
  k8s\_target\_port-misconfig-rca-2 & \pT & \pF & \pF & \pF & \pF & 1/5 \\
  k8s\_target\_port-misconfig-rca-3 & \pF & \pF & \pT & \pF & \pF & 1/5 \\
  misconfig\_app\_hotel\_res-rca-1 & \pF & \pF & \pF & \pT & \pF & 1/5 \\
  redeploy\_without\_PV-rca-1 & \pF & \pF & \pF & \pF & \pF & 0/5 \\
  revoke\_auth\_mongodb-rca-1 & \pF & \pF & \pF & \pF & \pF & 0/5 \\
  revoke\_auth\_mongodb-rca-2 & \pF & \pF & \pT & \pT & \pF & 2/5 \\
  scale\_pod\_zero\_social\_net-rca-1 & \pF & \pF & \pF & \pF & \pF & 0/5 \\
  user\_unregistered\_mongodb-rca-1 & \pF & \pF & \pF & \pF & \pF & 0/5 \\
  user\_unregistered\_mongodb-rca-2 & \pF & \pF & \pF & \pF & \pF & 0/5 \\
  wrong\_bin\_usage-rca-1 & \pF & \pF & \pF & \pF & \pF & 0/5 \\
  \midrule
  \textbf{Total} & 1/13 & 0/13 & 2/13 & 2/13 & 0/13 & -- \\
  \bottomrule
  \end{tabular}
  \end{table}

  \begin{table}[H]
  \centering
  \caption{Detailed results: \textbf{Mitigation} tasks (13 total).}
  \label{tab:detailed_mit}
  \scriptsize
  \setlength{\tabcolsep}{2pt}
  \begin{tabular}{lcccccc}
  \toprule
  \textbf{Problem ID} & \textbf{R1} & \textbf{R2} & \textbf{R3} & \textbf{R4} & \textbf{R5} & \textbf{Best} \\
  \midrule
  assign\_to\_non\_existent\_node-mit-1 & \pT & \pT & \pT & \pT & \pT & 5/5 \\
  auth\_miss\_mongodb-mit-1 & \pF & \pF & \pF & \pF & \pF & 0/5 \\
  k8s\_target\_port-misconfig-mit-1 & \pT & \pF & \pF & \pF & \pF & 1/5 \\
  k8s\_target\_port-misconfig-mit-2 & \pF & \pF & \pF & \pT & \pF & 1/5 \\
  k8s\_target\_port-misconfig-mit-3 & \pF & \pF & \pT & \pF & \pF & 1/5 \\
  misconfig\_app\_hotel\_res-mit-1 & \pF & \pF & \pF & \pF & \pF & 0/5 \\
  redeploy\_without\_PV-mit-1 & \pT & \pT & \pT & \pT & \pT & 5/5 \\
  revoke\_auth\_mongodb-mit-1 & \pF & \pF & \pF & \pF & \pF & 0/5 \\
  revoke\_auth\_mongodb-mit-2 & \pF & \pF & \pF & \pF & \pF & 0/5 \\
  scale\_pod\_zero\_social\_net-mit-1 & \pF & \pT & \pF & \pF & \pT & 2/5 \\
  user\_unregistered\_mongodb-mit-1 & \pF & \pF & \pF & \pF & \pF & 0/5 \\
  user\_unregistered\_mongodb-mit-2 & \pF & \pF & \pF & \pF & \pF & 0/5 \\
  wrong\_bin\_usage-mit-1 & \pF & \pF & \pF & \pF & \pF & 0/5 \\
  \midrule
  \textbf{Total} & 3/13 & 3/13 & 3/13 & 3/13 & 3/13 & -- \\
  \bottomrule
  \end{tabular}
  \end{table}